\documentclass[letterpaper]{article} 
\usepackage{aaai2026}  
\usepackage{subcaption}
\usepackage{times}  
\usepackage{helvet}  
\usepackage{courier}  
\usepackage[hyphens]{url}  
\usepackage{graphicx} 
\usepackage{natbib}  
\usepackage{caption} 
\usepackage[ruled,vlined,linesnumbered]{algorithm2e}
\SetAlgoCaptionLayout{centerline}
\SetKwInput{KwIn}{Input}
\SetKwInput{KwOut}{Output}
\SetKwInput{KwParam}{Parameter}
\SetAlCapFnt{\bfseries}
\SetAlCapNameFnt{\bfseries}
\SetAlgoNlRelativeSize{-1}
\SetAlgoNlRelativeSize{0}
\SetAlgoNlRelativeSize{0}
\DontPrintSemicolon
\usepackage{dsfont}

\usepackage{amsmath}
\usepackage{amsfonts}
\usepackage{soul, xcolor}

\usepackage{booktabs}
\usepackage{tabularx}

\usepackage{newfloat}
\usepackage{listings}
\usepackage[utf8]{inputenc}
\DeclareUnicodeCharacter{0327}{\c{}}

\DeclareCaptionStyle{ruled}{labelfont=normalfont,labelsep=colon,strut=off} 
\title{Interpret Policies in Deep Reinforcement Learning using SILVER with RL-Guided Labeling: A Model-level Approach to High-dimensional and Multi-action Environments}
\author {
	Yiyu Qian\textsuperscript{\rm 1}
    Liyuan Zhao\textsuperscript{\rm 2}
}
\affiliations {
    \textsuperscript{\rm 1}\{s4218854\}@rmit.edu.au,
    \textsuperscript{\rm 2}\{liyuanz6\}@uci.edu
}

\usepackage{bibentry}

\begin{document}

\maketitle

\begin{abstract}
Deep reinforcement learning (RL) achieves remarkable performance but lacks interpretability, limiting trust in policy behavior. The existing SILVER framework~\cite{li2025explainability} explains RL policy via Shapley-based regression but remains restricted to low-dimensional, binary-action domains. We propose \textbf{SILVER with RL-guided labeling}, an enhanced variant that extends SILVER to multi-action and high-dimensional environments by incorporating the RL policy’s own action outputs into the boundary points identification. Our method first extracts compact feature representations from image observations, performs SHAP-based feature attribution, and then employs RL-guided labeling to generate behaviorally consistent boundary datasets. Surrogate models, such as decision trees and regression-based functions, are subsequently trained to interpret RL policy’s decision structure. We evaluate the proposed framework on two Atari environments using three deep RL algorithms and conduct human-subject study to assess the clarity and trustworthiness of the derived interpretable policy. Results show that our approach maintains competitive task performance while substantially improving transparency and human understanding of agent behavior. This work advances explainable RL by transforming SILVER into a scalable and behavior-aware framework for interpreting deep RL agents in high-dimensional, multi-action settings.
\end{abstract}

\textbf{Code —} \url{https://github.com/qyy752457002/Interpret-DRL-using-SHAP-Project}

\begin{figure*}[t]
    \centering
    \includegraphics[width=0.7\textwidth]{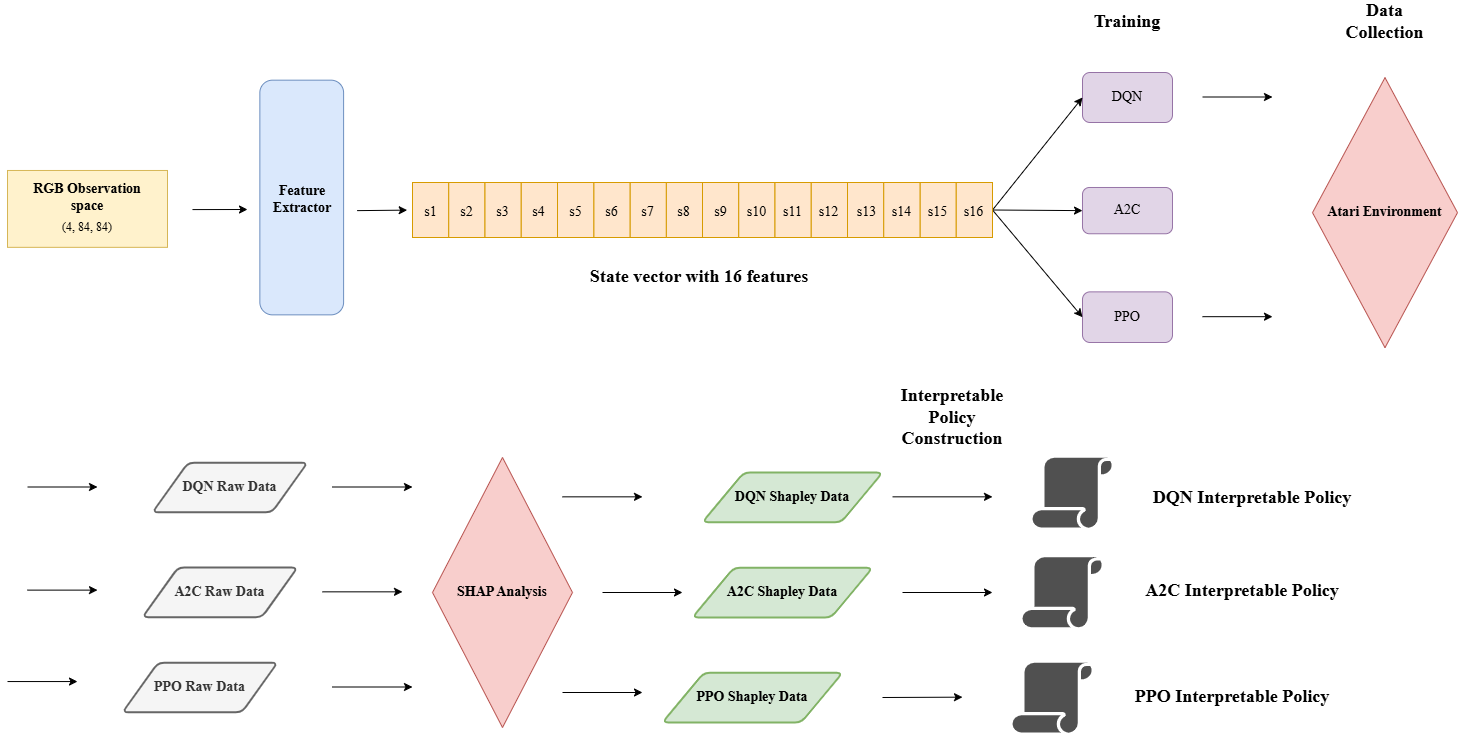}
    \caption{SILVER with RL-Guided Labeling: an overview of the proposed method illustrating the process of feature extraction, deep RL algorithms training, SHAP analysis, and interpretable policy construction}
    \label{fig:silver_method_overview}
\end{figure*}

\section{Introduction}
Deep reinforcement learning (RL) has achieved notable success across diverse domains, including strategic gameplay (e.g., AlphaGo Zero~\cite{shaheen2025reinforcement}), Atari games~\cite{jang2024analyzing}, robotics~\cite{tang2025deep}, and the fine-tuning of large language models (LLMs)~\cite{zhai2024fine}. These advances are largely attributable to the representational power of deep neural networks (DNNs), which excel at feature extraction and processing high-dimensional data. However, the opaque nature of DNNs poses significant challenges for understanding and interpreting the decision-making processes of RL agents~\cite{murad2024unraveling,fan2023rethinking}. This lack of transparency is particularly concerning in high-stakes applications~\cite{wang2024trustmis,shu2026trustworthy}, where interpretability is essential for ensuring safety and trust.

Explainable RL (XRL) has emerged to address similar challenges by enabling human users to understand the decision-making processes of RL agents. Within this context, XRL distinguishes between interpretability and explainability, each with a distinct focus. Interpretability refers to the intrinsic transparency of a model's structure and operation~\cite{li2025explainability}, often achieved through simpler models such as decision trees~\cite{costa2024evolving} or logistic regression~\cite{tahirovic2023interpretability}, which render the policy self-explanatory. Approaches based on this concept are typically categorized as model-level explanation methods~\cite{cheng2025survey}. Explainability, by contrast, involves post-hoc techniques that provide insights into the behavior of trained models, aiming to clarify or justify their decisions~\cite{li2025explainability}. These approaches include dataset-level explanation methods, which identify influential data in RL~\cite{cheng2025survey}. A prominent example is Shapley Additive Explanations (SHAP), which unifies various feature attribution methods within a theoretical framework grounded in Shapley values~\cite{lundberg2017unified}. Recent XRL research has applied SHAP to analyze the contribution of individual features in a state to the output of RL value functions~\cite{beechey2023explaining}~\cite{beechey2025theoretical}. SHAP provides local explanations by assigning numerical importance values to features in specific states~\cite{song2024comprehensive}, but it does not yield a global understanding of agent behavior.

To bridge the gap between interpretability and explainability, ~\cite{li2025explainability} introduced the SILVER (Shapley value-based Interpretable policy Via Explanation Regression) framework, which combines SHAP analysis with model-level explanation techniques to interpret deep RL policy. Although effective in relatively simple environments such as CartPole and MountainCar, SILVER relies on two key assumptions: first, that the observation space consists of discrete, well-defined features, which simplifies the computation of Shapley values; and second, that the policy operates within a binary-action setting. Both assumptions fail in complex domains with high-dimensional state spaces and large action spaces. In such environments, the absence of explicit feature definitions hinders direct SHAP analysis, while the exponential growth of pairwise action boundaries renders binary-action formulations inadequate. Prior XRL studies in high-dimensional domains have attempted to address similar challenges through saliency map visualizations for decision relevance~\cite{greydanus2018visualizing} or adapted versions of SHAP in applied contexts such as counter-drone operations~\cite{ccetin2024analyzing}. Yet, these methods either fail to produce globally interpretable policy or neglect to assess the explanatory effectiveness of their results. Consequently, there remains a pressing need for XRL approaches capable of handling multi-action, high-dimensional environments while providing both interpretability and explainability.

To address these limitations, we propose \textbf{SILVER with RL-guided labeling}, an enhanced framework that extends SILVER to multi-action and high-dimensional environments. Our approach integrates the RL policy's own action outputs into the boundary points identification, ensuring that the resulting decision boundaries accurately reflect the agent’s actual behavior rather than geometric proximity alone. This modification allows the construction of interpretable policy that remains consistent with RL policy, even in visually complex domains.

This paper makes three primary contributions:

\begin{itemize}
\item \textbf{Scalable Interpretability in High-Dimensional Environments}: We extend SILVER to operate on visually rich Atari domains, where both the state and action spaces are significantly more complex than in low-dimensional control tasks. This extension enables SHAP analysis on compact feature representations learned from raw RGB observations, enhancing the framework’s scalability and robustness.

\item \textbf{Interpretable Policy Derivation with RL-Guided Labeling}: We introduce a RL-guided labeling mechanism that uses the RL policy’s action outputs to annotate boundary points, enabling interpretable policy derivation in multi-action settings. We analyze performance of multiple interpretable models, such as decision trees, linear regression, and logistic regression.

\item \textbf{Human-Centered Evaluation of Comprehensibility and Trust}: We conduct human-subject study to evaluate how the derived interpretable policy supports user understanding and trust. Through structured feedback, we assess the extent to which these models enhance the perceived transparency and reliability of RL agents’ decision-making processes.
\end{itemize}


\section{Related Work}
This section reviews related work on XRL methods in explainability and interpretability.

\subsection{Explainability}
Explainability-focused XRL methods analyze how training data influence policy learning. The influence function~\cite{koh2017understanding} estimates the effect of reweighting a sample on model parameters and can be extended to RL for assessing replay transitions.

Shapley values~\cite{shapley1953value} provide a principled foundation for attribution.
Extensions such as Data Shapley~\cite{Ghorbani2019} and FreeShap~\cite{Wang2024} improve scalability and quantify each sample’s contribution to policy formation.

Recent studies move beyond static attribution toward behavioral and counterfactual reasoning: process-mining methods~\cite{qian2025exploring} expose sequential decision logic; SVERL~\cite{beechey2025sverl} unifies behavioral and predictive perspectives; and counterfactual explainers~\cite{dong2025counterfactual} illustrate how alternative actions change outcomes.

\subsection{Interpretability}
Interpretability-oriented XRL improves transparency by expressing policies in human-understandable forms~\cite{glanois2024survey}.

Direct methods design interpretable controllers such as decision trees~\cite{mahbooba2021explainable} or analytical formulas~\cite{hein2018interpretable}, with recent advances in differentiable trees via distillation~\cite{gokhale2024distill2explain}.

Indirect methods first train RL policies and then convert them into interpretable surrogates, e.g., PIRL~\cite{vouros2022explainable} and VIPER~\cite{saulieres2024explaining}.

Our approach follows the indirect paradigm, using Shapley-based RL-guided labeling to transform deep RL policies into compact surrogate models for high-dimensional, multi-action environments, ensuring transparency without sacrificing performance.

\section{Preliminaries}

\subsection{Foundation of Reinforcement Learning}
RL is a subﬁeld of machine learning that focuses on training agents to make sequential decisions by interacting with an environment~\cite{sutton2018reinforcement}. The environment is framed as a Markov Decision Process (MDP):

\begin{itemize}
    \item \(S\): The set of possible states of the environment.
    \item \(A\): The set of available actions for the agent.
    \item \(P(s' | s, a)\): The probability of transitioning from state \(s\) to state \(s'\) when action \(a\) is taken.
    \item \(R(s, a)\): The reward specifies the benefit or cost of executing a particular action \(a\) in state \(s\).
    \item \(\gamma \in (0, 1)\): The discount factor determines how much a future reward should be discounted compared to a current reward.
\end{itemize}

The goal of RL is to learn an optimal policy \(\pi(a|s)\) that maximizes the agent’s long-term reward.

\subsection{Shapley Values in Reinforcement Learning}

The Shapley value~\cite{das2022values} is a concept from cooperative game theory that allocates credit for the total value \( v(N) \) earned by a team \( N \) among its players. The value is defined as:

\begin{equation}
\label{eq:shapley-value}
\phi_i(v) = \sum_{C \subseteq N \setminus \{i\}} \frac{|C|!(n - |C| - 1)!}{(n!)} \left[v(C \cup \{i\}) - v(C)\right]
\end{equation}
where \( v(C) \) indicates the value generated by a coalition of players \( C \). The Shapley value \( \phi_i(v) \) represents the average marginal contribution of player \( i \) when added to all possible coalitions \( C \).

In RL, the state features \( \{s_1, ..., s_n\} \) can be treated as players, and the policy output \( \pi(s) \) is considered the total value generated by their contributions. To calculate the Shapley values of these features, it is necessary to define a characteristic function \( v(C) \) that reflects the model's output for a coalition of features \( s_C \subseteq \{s_1, ..., s_n\} \).

Given that the trained policy is undefined for partial inputs \( s_C \), it is important to properly define the characteristic function to ensure accurate Shapley values. Following the on-manifold characteristic value function approach~\cite{frye2020shapley}~\cite{beechey2023explaining}~\cite{beechey2025theoretical}, we account for correlations between features rather than assuming their independence.

For a deterministic policy \( \pi: S \rightarrow A \), which outputs actions, the characteristic function is given by:

\begin{equation}
\label{eq:char-deterministic}
v^{\pi}(C) := \pi_C(s) = \sum_{s' \in S} p^{\pi}(s'|s_C) \pi(s')
\end{equation}
where \( s' = s_C \cup s_C' \), and \( p^{\pi}(s'|s_C) \) is the probability of being in state \( s' \) given the partial state features \( s_C \) observed under policy \( \pi \). For a stochastic policy \( \pi: S \times A \rightarrow [0, 1] \), which outputs action probabilities, the characteristic function is:

\begin{equation}
\label{eq:char-stochastic}
v^{\pi}(C) := \pi_C(a|s) = \sum_{s' \in S} p^{\pi}(s'|s_C) \pi(a|s')
\end{equation}

\section{Methodology}
Figure~\ref{fig:silver_method_overview} illustrates the pipeline of our methodology. We trained deep RL models using Stable-Baselines3~\cite{raffin2021stable}, a widely adopted RL framework, and employed a CNN-based feature extractor to obtain compact representations from raw RGB observations. Using trained deep RL models, we collected data from 1,000 Atari episodes and conducted SHAP analysis on this data to generate Shapley vectors that capture feature contributions to the agent’s decision-making. Based on these Shapley vectors, we constructed interpretable policy and subsequently conducted human-subject study to evaluate the comprehensibility and trustworthiness of the derived policy. 

\subsection{Feature Extraction}
Figure~\ref{fig: Features Extractor} shows our CNN-based feature extractor, which converts stacked RGB Atari frames $(4,84,84)$ into a compact state vector of 16 features. Such low-dimensional representations are critical for achieving stable and efficient policy learning, as they distill high-dimensional visual inputs into task-relevant features while minimizing redundancy~\cite{echchahed2025survey}. This design is fully compatible with Stable-Baselines3, using the NatureCNN encoder within the CnnPolicy architecture.

\subsection{SHAP Analysis}
Following the SILVER framework, we compute each feature’s attribution $\phi_i(v^\pi)$—the marginal contribution of feature $i$ in state $s$ under policy $\pi$. Specifically, we plug Eq.~\eqref{eq:char-deterministic} or Eq.~\eqref{eq:char-stochastic} into the Shapley definition in Eq.~\eqref{eq:shapley-value}.  

These attributions indicate how features steer the policy’s choice. For example, in a two-action setting ($a_1=-1$, $a_2=1$), a positive $\phi_i(v^\pi)$ suggests feature $i$ pushes the policy toward $a_2$, while a negative value pulls it toward $a_1$. Because features with identical marginal effects receive the same score, symmetries in policy reasoning are directly revealed. Finally, we collect all feature attributions into the vector
\[
  \Phi_s = (\phi_1, \dots, \phi_n),
\]
and cluster these vectors to identify groups of states that induce similar action preferences, thereby further gives insights into action-group boundaries.

\subsection{Interpretable Policy Construction}
This section describes how we construct interpretable policy. The complete algorithm is presented in Algorithm~\ref{alg:silver_rl_Guided}. It consists of four stages:

\begin{itemize}
    \item \textbf{Action K-Means Clustering}: 
    Group Shapley vectors into $|A|$ clusters corresponding to the agent’s discrete actions.
    
    \item \textbf{Boundary Points Identification}: 
    Locate boundary points between action clusters.
    
    \item \textbf{Inverse Mapping and RL-Guided Labeling}: 
    Map boundary points back to their corresponding states and query the RL policy to assign action labels to each state.
    
    \item \textbf{Interpretable Models Fitting}: 
    Fit decision trees, linear regression, and logistic regression to approximate decision boundaries.
\end{itemize}

\subsubsection{Action K-Means Clustering}
For each state $s$, we compute its Shapley vector $\Phi_s \in \mathbb{R}^n$, where each entry quantifies the marginal contribution of a feature. We then perform K-means clustering with $k = |A|$ clusters, equal to the number of discrete actions:
\begin{equation}
\arg\min_{A} \sum_{i=1}^{k} \sum_{\Phi_s \in A_i} \| \Phi_s - \mu_i \|^2,
\end{equation}
where $\mu_i$ is the centroid of cluster $A_i$:
\[
\mu_i = \frac{1}{|A_i|} \sum_{\Phi_s \in A_i} \Phi_s.
\]
This ensures that each action is associated with a representative cluster of Shapley vectors.

\subsubsection{Boundary Points Identification}
After forming clusters, boundaries between different action regions can be identified through the use of boundary points. 
A \textit{boundary point} $X$ lies at the interface of two clusters $A_i$ and $A_j$, 
where the policy exhibits equal likelihood of choosing either action. 
This condition reflects uncertainty in the policy’s action selection at a given state, 
and thus $X$ represents a critical decision boundary. 
Formally, the boundary point is determined by minimizing the difference between its distances to the two cluster centroids:
\begin{equation}
    \arg\min_X \left( \|X - \mu_i\|^2 - \|X - \mu_j\|^2 \right),
    \label{eq:boundary_minimization}
\end{equation}
where $\mu_i$ and $\mu_j$ denote the centroids of points in $A_i$ and $A_j$, respectively.

\subsubsection{Inverse Mapping and RL-Guided Labeling}
Once boundary points are identified in the Shapley vector space, they must be mapped back to their corresponding states in the original environment. We follow the \textit{Inverse Shapley Mapping} defined by~\cite{li2025explainability}:
\[
\phi_i^{-1} : \phi_i(v) \rightarrow \{i\},
\] 
which recovers the original state $s$ associated with a given Shapley vector $\Phi_s$. 

The original SILVER framework was designed for binary-action environments. It assumes that states with similar Shapley vectors correspond to the same action, allowing decision boundaries to be inferred purely from distances between action clusters in the Shapley vector space. This assumption holds in binary-action settings ($a_1 = left$ and $a_2 = right$), where each boundary point naturally lies between two well-defined action clusters. In multi-action environments, however, this geometric assumption breaks down. A single state may lie near several action clusters in the Shapley vector space, making it unclear which action the boundary point truly represents. As the number of actions increases, the number of possible action pairs grows rapidly, and distances in Shapley vector space no longer provide reliable information about RL agent’s actual decision-making behavior. Consequently, boundary points defined solely by geometric proximity become ambiguous and cannot be used to train accurate interpretable models.

To resolve this problem, we introduce a \textit{RL-guided labeling mechanism}. Table~\ref{alg:silver_rl_Guided} shows the comparison between SILVER and SILVER with RL-guided labeling. 

After reconstructing each boundary state $s_{ij}$ through inverse mapping, we query the trained RL policy $\pi$ to obtain the corresponding action:
\[
a_{ij} = \pi(s_{ij}).
\]
This step ensures that every boundary point receives an action label derived directly from the RL policy. By combining inverse mapping with RL-guided labeling, we obtain a behaviorally consistent boundary dataset $(s_{ij}, a_{ij})$, which enables reliable training of interpretable models that approximate decision boundaries.

\subsubsection{Interpretable Models Fitting}
For labeled boundary states $\{(s_{ij}, a_{ij})\}$, we fit interpretable models. Unlike~\cite{li2025explainability}, which employed linear regression exclusively, we explore multiple interpretable models to capture different aspects of policy behavior. Specifically, decision tree exposes hierarchical if–then reasoning patterns, linear regression yields direct coefficient-based explanations, and logistic regression represents softmax probabilities among actions.

\subsection{Human Study}
To assess the interpretability and trustworthiness of our framework, we conducted a human-subject study, following recent XRL evaluation standards~\cite{kohler2025evaluating}~\cite{alabdulkarim2025experiential}. The study consists of two tasks designed to measure participants’ ability to comprehend and trust interpretable policy.

\subsubsection{Policy Comprehension Task} 
Participants are presented with selected states from Atari environments, together with the corresponding interpretable policy. They are asked to manually infer which action the agent would take based on interpretable policy, following the policy interpretability protocol proposed by~\cite{kohler2025evaluating}. This task evaluates how effectively humans can understand and predict the agent’s actions based on interpretable policy.

\subsubsection{Trust Calibration Task} 
Participants are presented with paired gameplay clips of agents acting under either the original policy or a fixed interpretable policy. They are asked to rate how much they trust each agent’s decisions using a 5-point Likert scale, adapted from the interpretability evaluation survey implemented in~\cite{sarch2025grounded}. This task assesses how reliably and consistently humans perceive the surrogate policy compared to the original policy.

\section{Experiments}
To evaluate the effectiveness of SILVER with RL-guided labeling, we performed experiments. To assess scalability, we applied our framework to high-dimensional RGB environments from the Atari benchmark, which presents substantially more complex state and action spaces than classical control tasks. To derive interpretable policy, we employed classification-based and regression-based techniques, including decision trees, linear regression, and logistic regression, thereby enabling an examination of the trade-offs between model interpretability and policy performance. To test real-world applicability, we conducted human-subject study to evaluate the comprehensibility and trustworthiness of the derived policy. 

All Atari environments were wrapped using the \texttt{Atari Wrappers} from Stable-Baselines3, which clip rewards to $[-1, 0, 1]$ by its sign for training stability. 

The experimental results demonstrate that SILVER with RL-guided labeling scales effectively to complex environments, produces interpretable policy that balances transparency with performance, and enhances user trust and confidence in the decision-making of RL agents.

\begin{figure*}[!h]
    \centering
    \begin{subfigure}[t]{0.33\textwidth}
        \centering
        \includegraphics[width=\linewidth]{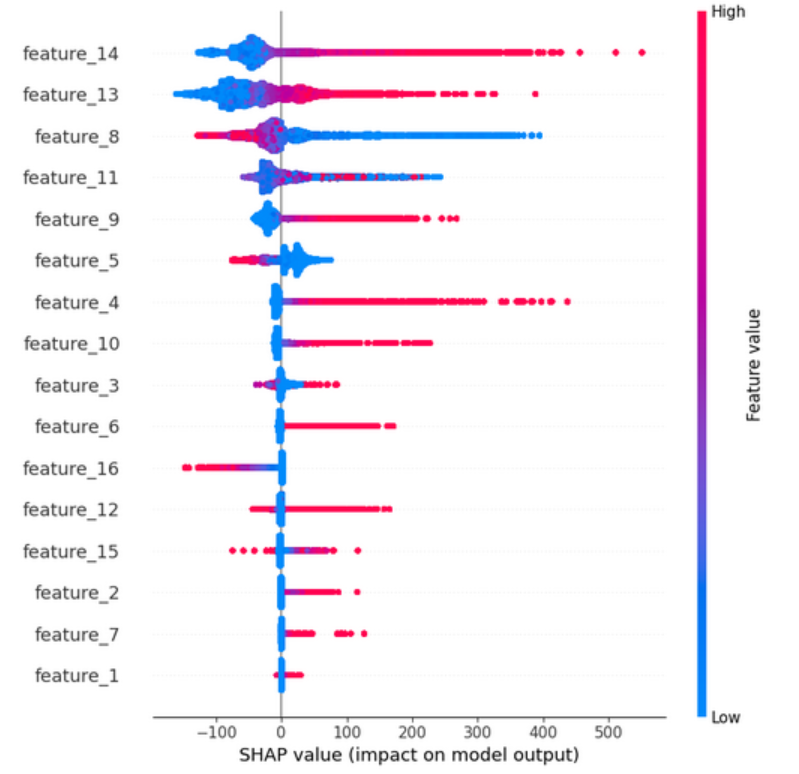}
        \caption{MsPacman DQN - SHAP value}
        \label{fig:MsPacman_DQN_SHAP}
    \end{subfigure}
    \hfill
    \begin{subfigure}[t]{0.33\textwidth}
        \centering
        \includegraphics[width=\linewidth]{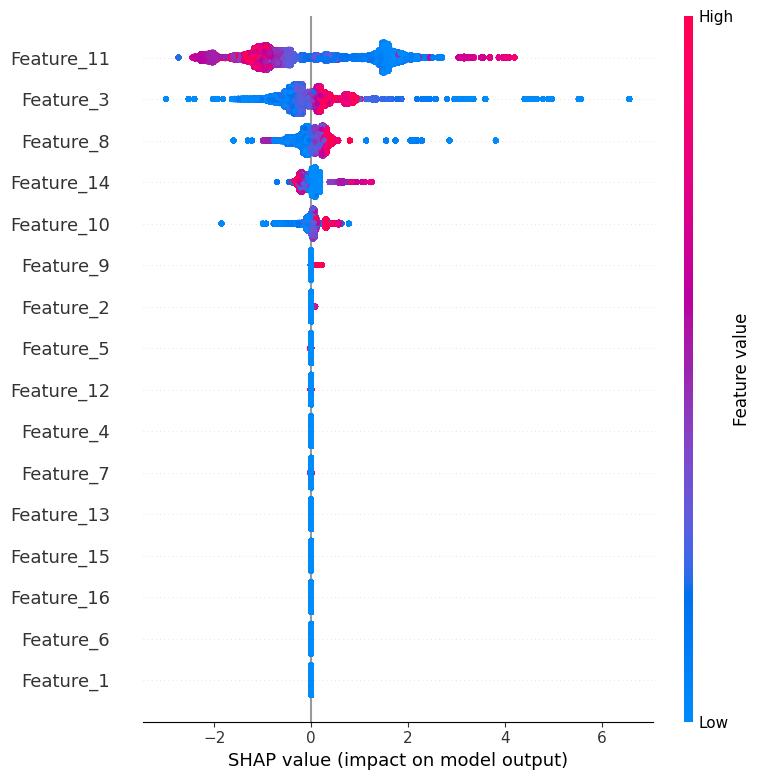}
        \caption{MsPacman A2C - SHAP value}
        \label{fig:MsPacman_A2C_SHAP}
    \end{subfigure}
    \hfill
    \begin{subfigure}[t]{0.33\textwidth}
        \centering
        \includegraphics[width=\linewidth]{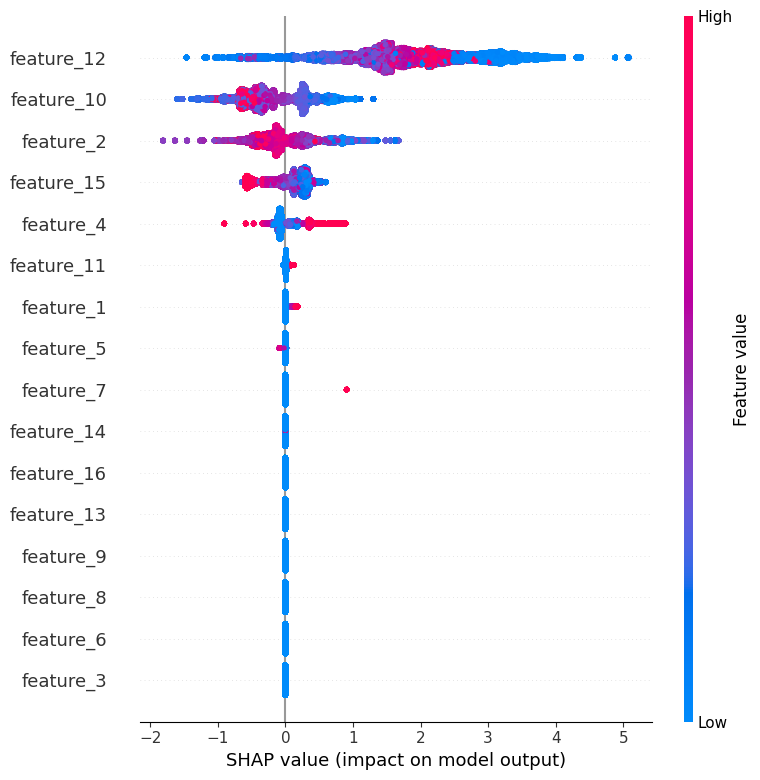}
        \caption{MsPacman PPO - SHAP value}
        \label{fig:MsPacman_PPO_SHAP}
    \end{subfigure}
    \caption{SHAP value comparison of MsPacman under different algorithms (DQN, A2C, PPO)}
    \label{fig:MsPacman_SHAP_comparison}
\end{figure*}

\begin{figure*}[!h]
    \centering
    \begin{subfigure}[t]{0.33\textwidth}
        \centering
        \includegraphics[width=\linewidth]{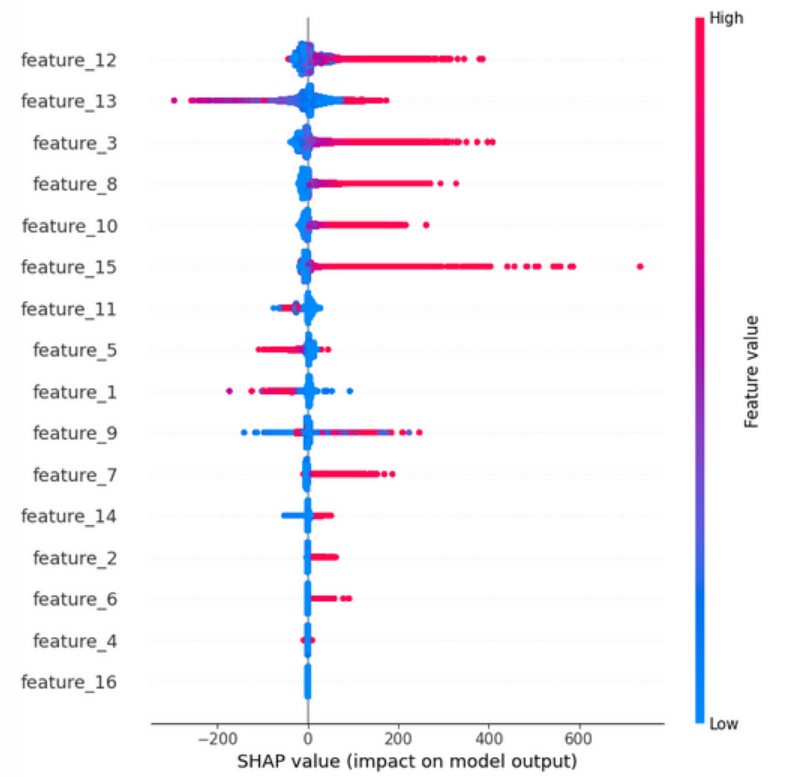}
        \caption{RoadRunner DQN - SHAP value}
        \label{fig:RoadRunner_DQN_SHAP}
    \end{subfigure}
    \hfill
    \begin{subfigure}[t]{0.33\textwidth}
        \centering
        \includegraphics[width=\linewidth]{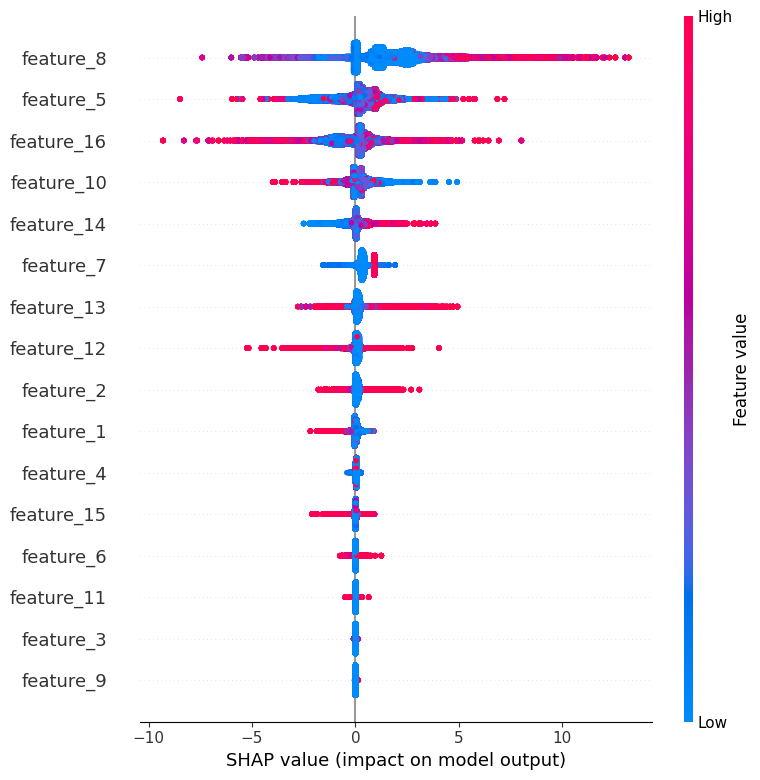}
        \caption{RoadRunner A2C - SHAP value}
        \label{fig:RoadRunner_A2C_SHAP}
    \end{subfigure}
    \hfill
    \begin{subfigure}[t]{0.33\textwidth}
        \centering
        \includegraphics[width=\linewidth]{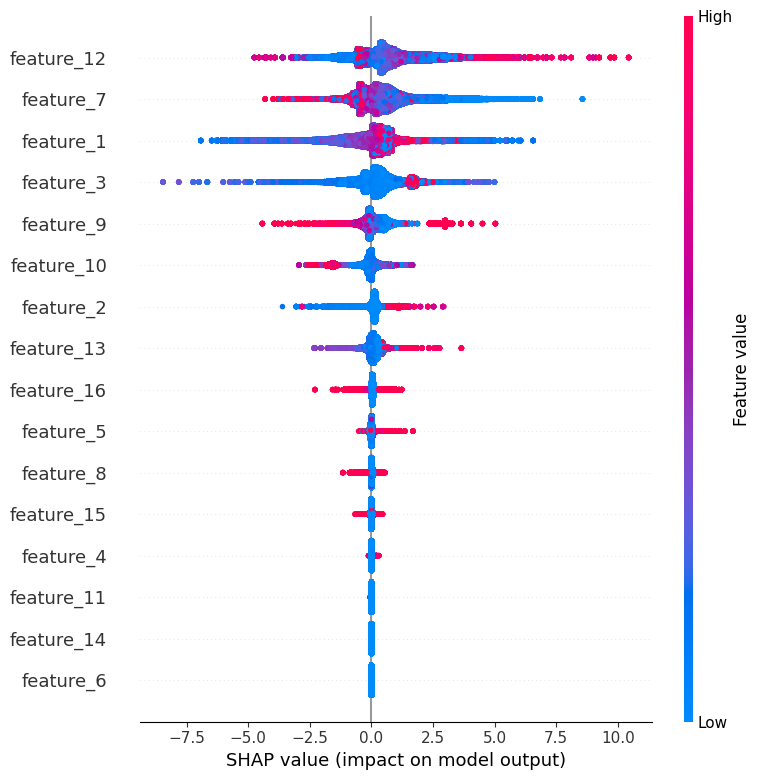}
        \caption{RoadRunner PPO - SHAP value}
        \label{fig:RoadRunner_PPO_SHAP}
    \end{subfigure}
    \caption{SHAP value comparison of RoadRunner under different algorithms (DQN, A2C, PPO)}
    \label{fig:RoadRunner_SHAP_comparison}
\end{figure*}

\begin{figure*}[!h]
    \centering
    \begin{subfigure}[b]{0.48\textwidth}
        \centering
        \includegraphics[width=\linewidth]{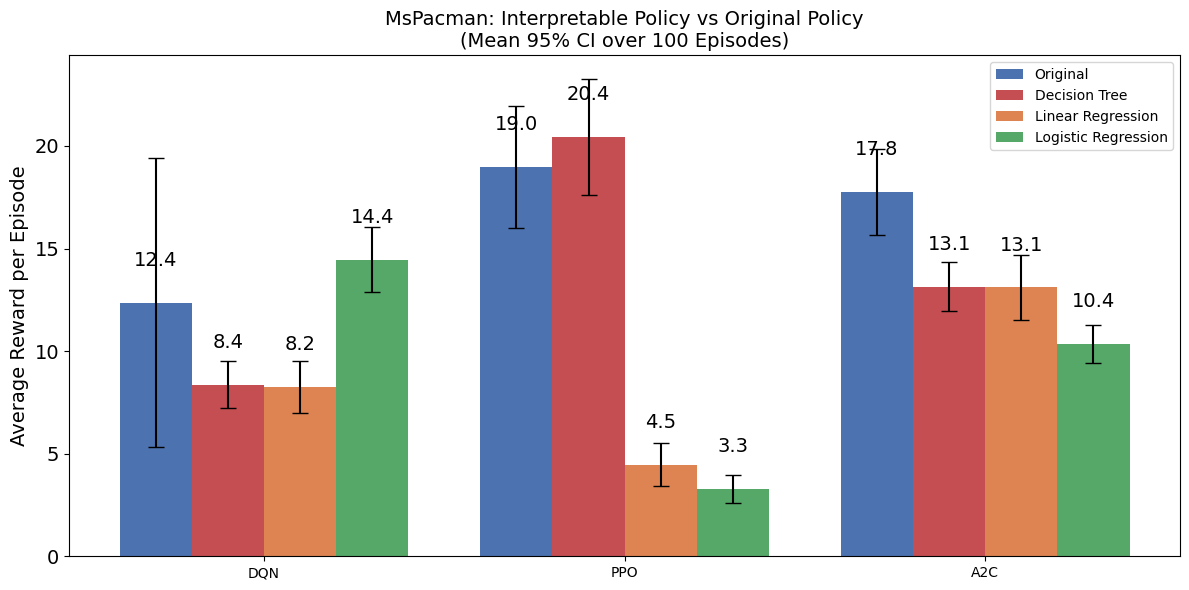}
        \caption{MsPacman: interpretable policy vs original policy (mean 95\% CI over 100 episodes)}
        \label{fig:MsPacman_Mean_95_CI}
    \end{subfigure}
    \hfill
    \begin{subfigure}[b]{0.48\textwidth}
        \centering
        \includegraphics[width=\linewidth]{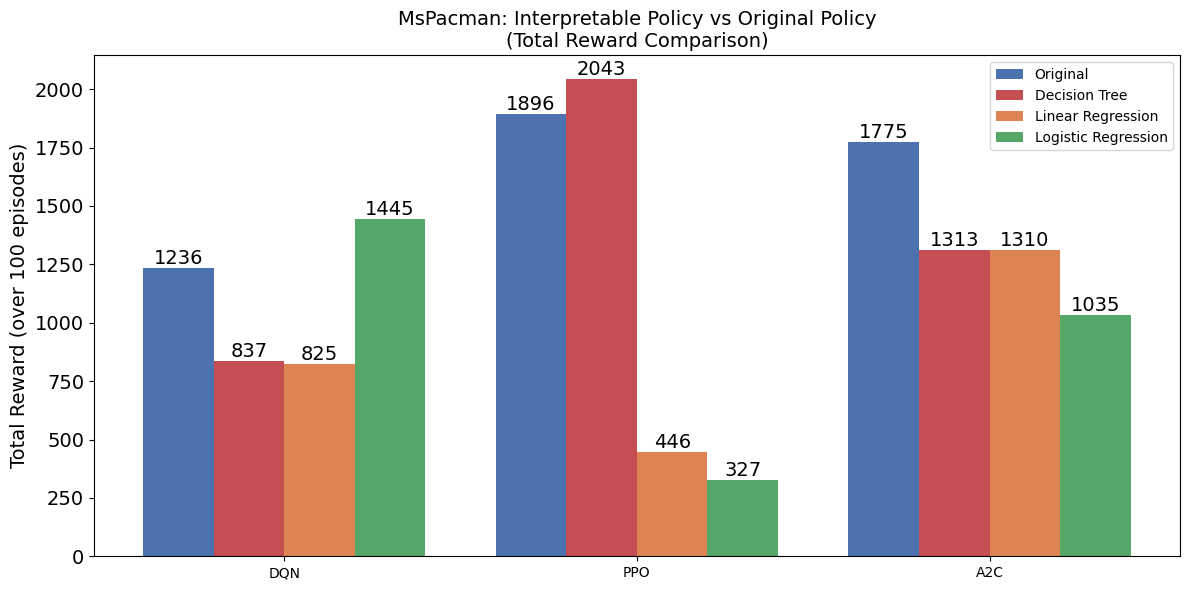}
        \caption{MsPacman: interpretable policy vs original policy (total reward comparison)}
        \label{fig:MsPacman_Total_Reward}
    \end{subfigure}
    \caption{Performance comparison between interpretable and original policies in MsPacman}
    \label{fig:MsPacman_Performance_Comparison}
\end{figure*}

\begin{figure*}[!h]
    \centering
    \begin{subfigure}[b]{0.48\textwidth}
        \centering
        \includegraphics[width=\linewidth]{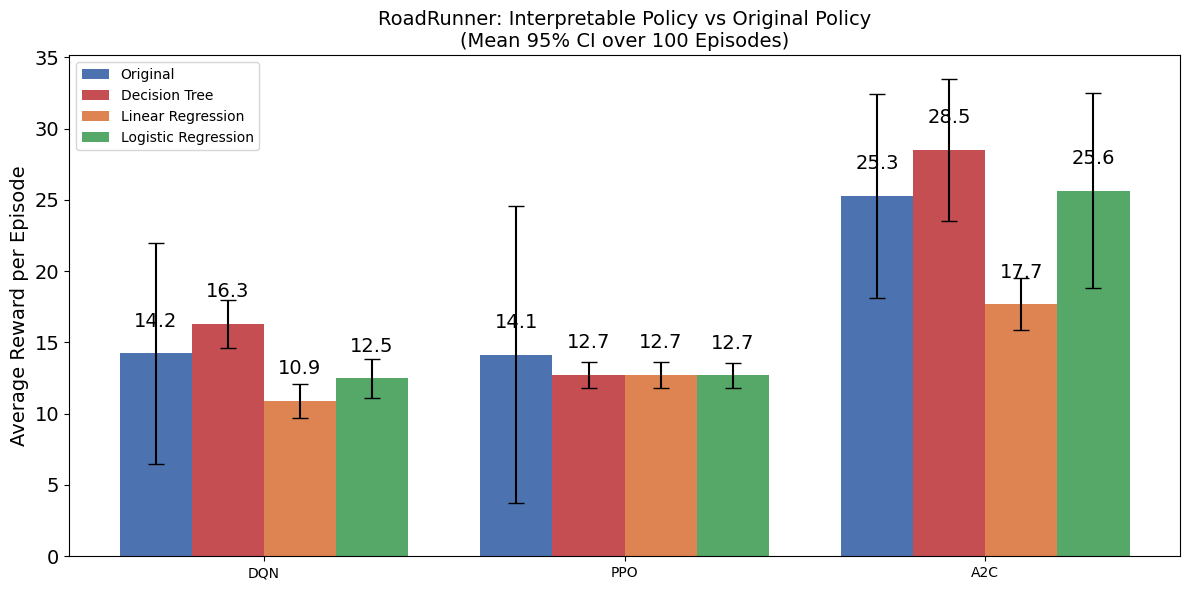}
        \caption{RoadRunner: interpretable policy vs original policy (mean 95\% CI over 100 episodes)}
        \label{fig:RoadRunner_Mean_95_CI}
    \end{subfigure}
    \hfill
    \begin{subfigure}[b]{0.48\textwidth}
        \centering
        \includegraphics[width=\linewidth]{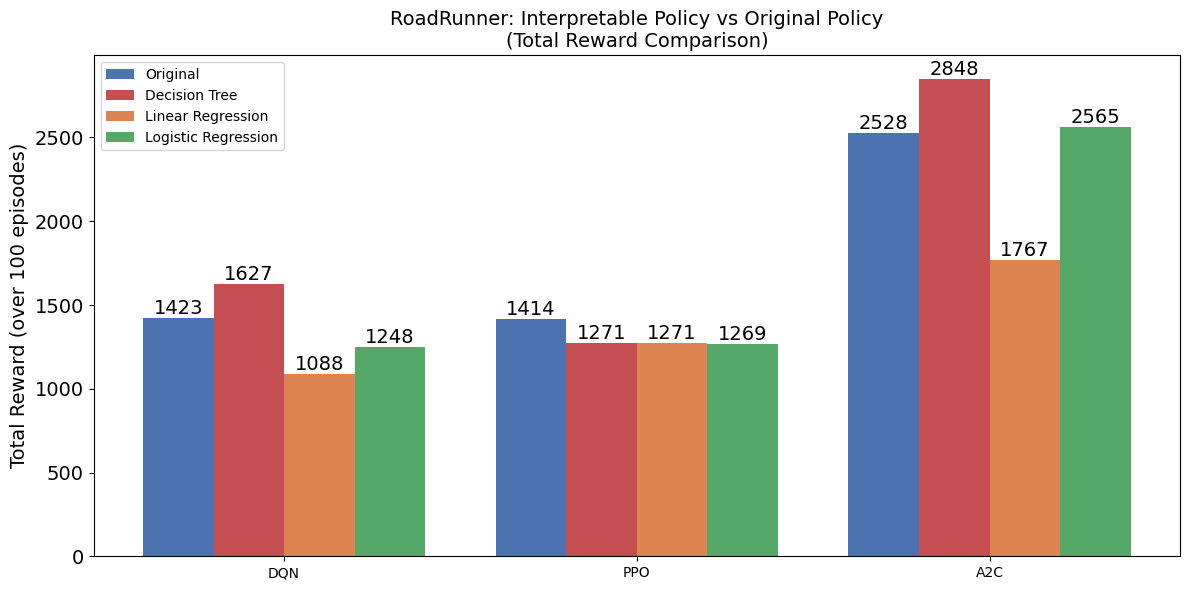}
        \caption{RoadRunner: interpretable policy vs original policy (total reward comparison)}
        \label{fig:RoadRunner_Total_Reward}
    \end{subfigure}
    \caption{Performance comparison between interpretable and original policies in RoadRunner}
    \label{fig:RoadRunner_Performance_Comparison}
\end{figure*}

\subsection{MsPacman}
The MsPacman environment is a maze-based Atari game where the agent navigates to collect rewards and avoid ghosts. The observation space consists of high-dimensional RGB frames showing MsPacman, pellets, fruits, and ghosts. The action space includes 9 discrete movements with or without the fire button. The agent gains rewards for consuming pellets and fruits, and receives a penalty when caught by a ghost. The goal is to maximize the cumulative score through strategic navigation.

We sampled state distributions from 1000 episodes collected by each deep RL algorithm. For every sampled state, we computed the Shapley values of its features using Equation (5), thereby obtaining a Shapley vector that quantifies the contribution of individual features to the policy’s decision. Figures~\ref{fig:MsPacman_SHAP_comparison} presents the Shapley values of 16 features for DQN, PPO, and A2C, respectively.

Based on these Shapley values, we applied k-means clustering to the action space in order to identify cluster centroids, with each cluster corresponding to a distinct action region. Since MsPacman comprises 9 discrete actions, we formed 9 clusters and computed boundary points between them, yielding 36 boundary points in total.

Subsequently, we reconstructed boundary points in the original state space by inverting Shapley values. Tables \ref{tab:MsPacman_DQN_boundary_points}, \ref{tab:MsPacman_DQN_shap_inverse_boundary_points}, \ref{tab:MsPacman_A2C_boundary_points}, \ref{tab:MsPacman_A2C_shap_inverse_boundary_points}, \ref{tab:MsPacman_PPO_boundary_points}, \ref{tab:MsPacman_PPO_shap_inverse_boundary_points} report the boundary points and their Shapley-inverse counterparts for each algorithm. Finally, we applied decision trees, linear regression, and logistic regression to obtain interpretable policy for DQN, PPO, and A2C.

\paragraph{Decision Tree} 
Figures~\ref{fig:MsPacman_ppo_decision_tree}, \ref{fig:MsPacman_a2c_decision_tree}, and \ref{fig:MsPacman_dqn_decision_tree} illustrate decision trees for PPO, A2C, and DQN, respectively. 

Taking PPO as an example, consider the first row in Table~\ref{tab:MsPacman_PPO_shap_inverse_boundary_points}. Starting from the root node of Figure~\ref{fig:MsPacman_ppo_decision_tree}, the decision process proceeds as follows:

\begin{enumerate}
  \item At the root node, check $Feature_{12} \leq 14173.283$. \\
        Since $8493.874 < 14173.283$, move to the \textbf{left child}.
  \item Next, check $Feature_{4} \leq 2304.729$. \\
        Since $530.9578 < 2304.729$, move to the \textbf{left child}.
  \item Then, check $Feature_{2} \leq 6694.608$. \\
        Since $5753.206 < 6694.608$, move to the \textbf{left child} and reach a leaf: 
        $\boxed{\text{class} = 7}$, which corresponds to  $\boxed{\text{action} =  7}$.
\end{enumerate}

\paragraph{Linear Regression}
Table~\ref{tab:MsPacman Linear Regression} illustrates the interpretable policy boundaries derived from linear regression for PPO, A2C, and DQN. Here, $x_{1}$ through $x_{16}$ represent the 16 state features used in the analysis.

Taking PPO as an example, consider the second row in table~\ref{tab:MsPacman_PPO_shap_inverse_boundary_points}. Substituting the feature values into the decision boundary in table~\ref{tab:MsPacman Linear Regression}, we obtain:

\[
\begin{aligned}
f(x) &= 7.37 \\
&\quad + (2.77 \times 10^{-4}) \times 6628.804 \\
&\quad - (1.94 \times 10^{-4}) \times 1914.645 \\
&\quad - (5.66 \times 10^{-4}) \times 1880.883 \\
&\quad + (1.29 \times 10^{-5}) \times 14349.77 \\
&\quad - (1.35 \times 10^{-4}) \times 11766.16 \\
&\approx 6.37.
\end{aligned}
\]

Taking the floor of this value yields $\boxed{f(x) = 6}$, which corresponds to $\boxed{\text{action} = 6}$.

Although actions are discrete, linear regression serves as a continuous decision boundary function. Its outputs are discretized (via flooring) to recover action indices, yielding a concise and interpretable mapping from state features to actions.

\paragraph{Logistic Regression}
Table~\ref{tab:MsPacman_A2C_LogReg}, Table~\ref{tab:MsPacman_PPO_LogReg}, and Table~\ref{tab:MsPacman_DQN_LogReg} illustrate the interpretable policy boundaries derived from logistic regression for A2C, PPO, and DQN.

Taking PPO as an example, consider the third row in Table~\ref{tab:MsPacman_PPO_shap_inverse_boundary_points}. Substituting the feature values into the corresponding decision boundaries in Table~\ref{tab:MsPacman_PPO_LogReg}, we obtain the following logits:

\[
\begin{aligned}
f_{5}(x) &\approx -8.93, \quad f_{6}(x) \approx 26.12, \\
f_{7}(x) &\approx -33.03, \quad f_{8}(x) \approx 16.12.
\end{aligned}
\]

Applying the softmax function to these logits converts them into normalized probabilities:
\[
P(a=i \mid x) = \frac{e^{f_i(x)}}{\sum_{j} e^{f_j(x)}}.
\]
The predicted action is then determined by selecting the one with the highest probability:
\[
\hat{a} = \arg\max_{i} P(a=i \mid x).
\]
For this state, the highest probability corresponds to $\boxed{f_{6}(x)}$, indicating that the model predicts $\boxed{\text{action} = 6}$.

\subsection{RoadRunner}
We also evaluated our framework on RoadRunner. Details of the experiment are provided in Appendix D.

\subsection{Fidelity Score}
Fidelity measures how well the surrogate policy covers the original policy~\cite{altmann2025surrogate}. The fidelity function proposed by~\cite{li2025explainability} further quantifies the behavioral difference between the interpretable policy and the original policy:

\begin{equation}
F(\pi_{\text{interp}}, \pi_{\text{orig}}) =
\frac{1}{|S|} \sum_{s \in S} \mathds{1}\{\pi_{\text{interp}}(s) = \pi_{\text{orig}}(s)\},
\label{eq:fidelity}
\end{equation}
where the original policy is treated as the ground truth.

Building on this notion, we define our fidelity score explicitly as the percentage of states where the interpretable and original policies output the same action. 

\begin{figure*}[!h]
    \centering
    \begin{subfigure}[t]{0.37\textwidth}
        \centering
        \includegraphics[width=\linewidth]{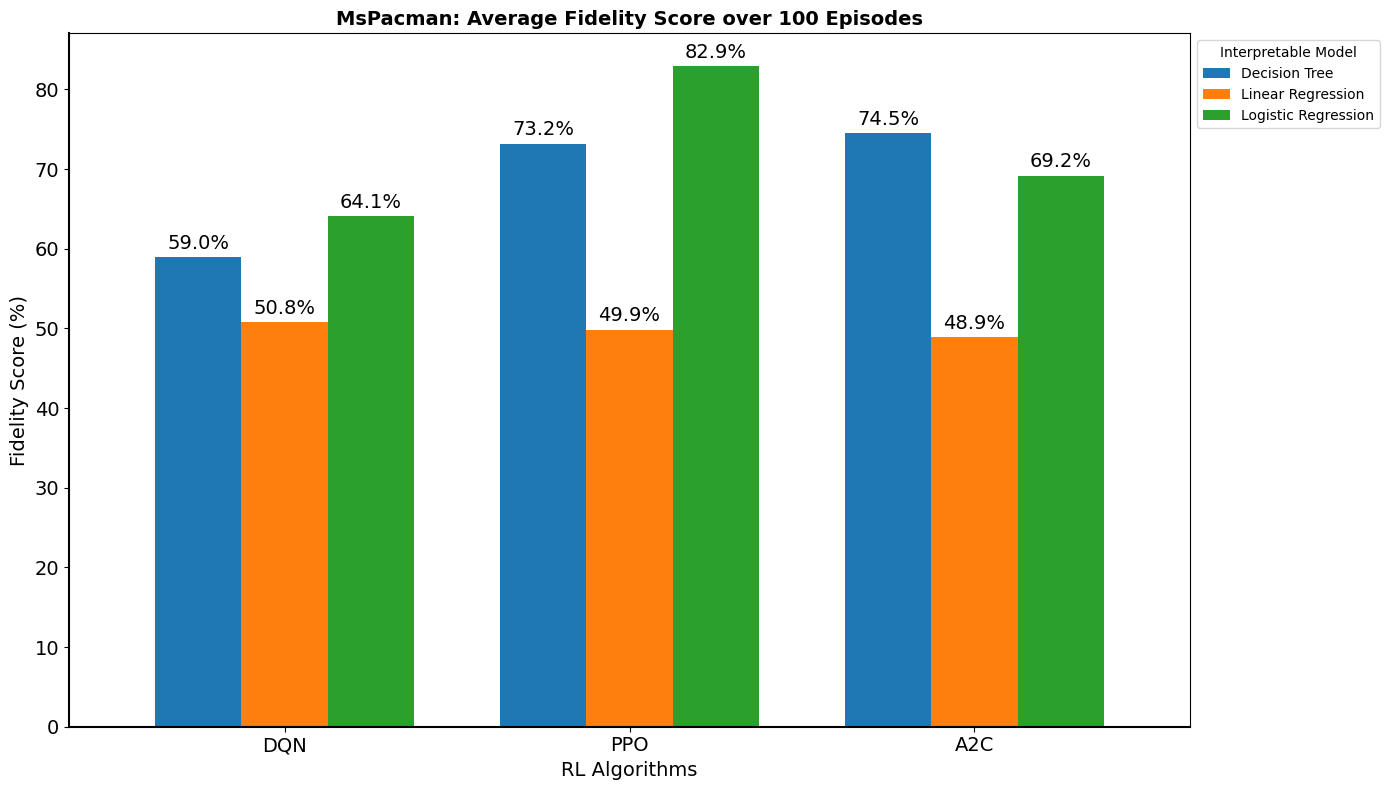}
        \caption{MsPacman: average fidelity score over 100 episodes}
        \label{fig:MsPacman_Fidelity_Score}
    \end{subfigure}
    \hspace{0.02\textwidth} 
    \begin{subfigure}[t]{0.37\textwidth}
        \centering
        \includegraphics[width=\linewidth]{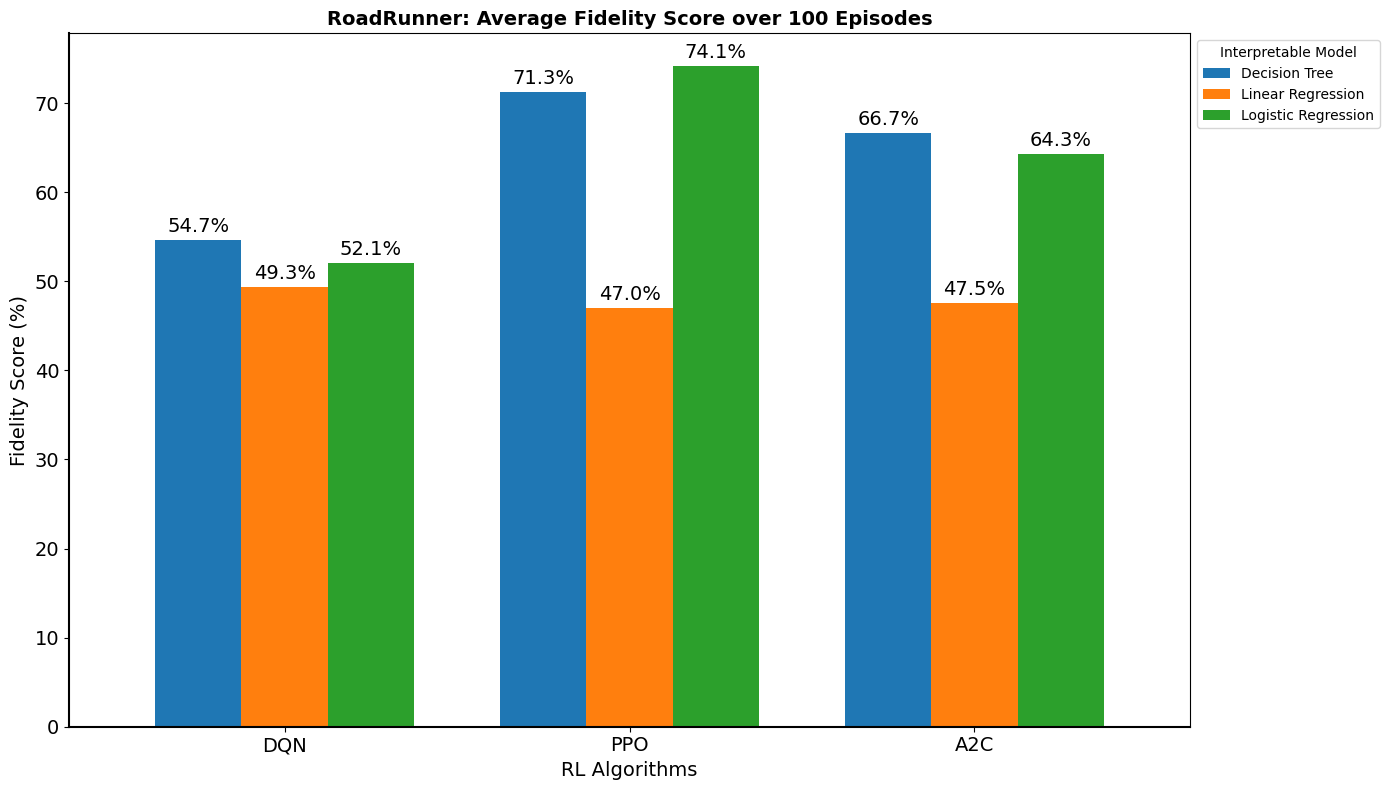}
        \caption{RoadRunner: average fidelity score over 100 episodes}
        \label{fig:RoadRunner_Fidelity_Score}
    \end{subfigure}
    \caption{Average fidelity scores of interpretable policies across MsPacman and RoadRunner}
    \label{fig:Fidelity_Score_Comparison}
\end{figure*}

Figure~\ref{fig:Fidelity_Score_Comparison} shows the fidelity scores across all environments and algorithms. 

\subsection{Human Study}
We conducted a human-subject study with 30 participants who had prior training in RL. To simplify evaluation and minimize cognitive burden, the study was restricted to the MsPacman environment and sole interpretable model. Since decision tree outperformed the PPO policy and provided the clearest if–then style rules, we selected it for evaluation.

\paragraph{Design}
Participants were presented with representative MsPacman states, together with the decision tree. Their task was to infer the action chosen by the decision tree. While doing so, we collected their action inferences, response times, and subjective trust ratings.

\paragraph{Metrics}
We measured three aspects:
\begin{itemize}
    \item \textbf{Response Time}: The average time taken to make an action inference, reflecting the efficiency of interpretability.
    \item \textbf{Accuracy}: The proportion of participant-inferred actions that matched the actual decision-tree outputs, reflecting the correctness of human interpretation.
    \item \textbf{Trust Ratings}: After completing all 4 trials, participants rated their overall trust in the decision-tree surrogate on a 5-point Likert scale, where:
    \begin{itemize}
        \item 1 = Very Low Trust (I would not rely on this model’s decisions at all)
        \item 2 = Low Trust (I have limited confidence; decisions often appear unreliable)
        \item 3 = Neutral (I am undecided; the model’s decisions sometimes make sense, sometimes not)
        \item 4 = High Trust (I generally have confidence in the model’s decisions, though not absolute)
        \item 5 = Very High Trust (I would fully rely on this model’s decisions with strong confidence)
    \end{itemize}
\end{itemize}

\section{Discussion}

\paragraph{MsPacman}
Decision tree derived from PPO slightly outperformed the original policy, whereas linear regression and logistic regression performed noticeably worse. For DQN and A2C, surrogate models were marginally below their RL baselines. Decision trees generally achieved the best and linear regression the weakest performance. An exception appeared in DQN, where logistic regression surpassed the decision tree. SHAP analyses revealed that high-index features ($>10$) contributed most to policy outputs: feature 12 in PPO, features 13, 14 in DQN, and feature 11 in A2C.

\paragraph{RoadRunner}
Decision trees from DQN and A2C slightly outperformed the original algorithms, while logistic regression under A2C achieved comparable results. Overall, decision trees remained strongest and linear regression weakest, except in PPO where all interpretable models performed similarly but below the original. Unlike MsPacman, features with lower indices also had strong influence, such as feature 8 in A2C and feature 3 in DQN.

\paragraph{Fidelity Score}
A high fidelity score does not necessarily imply strong task performance. For example, PPO’s logistic regression in MsPacman had high fidelity (82.9\%) but low performance, whereas the DQN decision tree in RoadRunner achieved superior performance despite low fidelity (54.7\%). Hence, fidelity reflects alignment between surrogate and RL policies but not their task-level effectiveness.

\section{Limitation}
Our work has three limitations that may need to be addressed in the future work:

\paragraph{Loss of Semantic Meaning in Feature Extraction}
The extracted state features lack explicit semantic meanings. Unlike classic control environments (e.g., CartPole, MountainCar) with well-defined variables, our state vectors lack such semantic clarity. We plan to integrate Vision Language Models (VLMs)~\cite{chen2024vision} to derive semantically grounded state features from complex, high-dimensional observations.

\paragraph{Limited to Atari Environments}
Although the framework has been evaluated in complex Atari domains, these environments do not fully represent real-world decision-making scenarios. With the growing interest in explainable RL fine-tuning for LLMs, extending the framework to transformer-based deep RL algorithms (e.g., GRPO, DAPO, GSPO) represents a promising future direction. In such contexts, user prompts could be modeled as observations, while the LLM’s output tokens serve as actions.

\paragraph{Restricted Scope of Human-subject Study}
The human-subject experiment was conducted on a single environment with one interpretable policy. Future work could involve broader evaluations across multiple environments and diverse interpretable policy forms to better assess generalizability and human comprehension.

\section{Conclusion}
In this paper, we propose SILVER with RL-guided labeling, a framework designed to generate interpretable policies for deep RL algorithms. Experimental results demonstrate that our method achieves performance comparable to deep RL algorithms in high-dimensional and multi-action environments, while preserving human comprehensibility and trust.

Potential future work includes: (1) refining the feature extraction process to assign semantic meaning to each state feature, (2) extending the framework to LLMs environments, particularly in the context of RL fine-tuning, and (3) conducting human-subject studies with diverse interpretable models to systematically compare their performance on comprehension and trust, and (4) deploying the compact surrogates under privacy-preserving secure inference~\cite{song2026oblivigate,wang2025privgnn}, where their low model complexity can substantially reduce cryptographic overhead compared with deep policies.


\bibliography{aaai2026}

\newpage

\appendix
\clearpage

\section*{\centering Appendix A: SILVER with RL-Guided Labeling} \label{Appendix A: SILVER with RL-Guided Labeling}

\begin{algorithm}[h]
\caption{\textbf{SILVER with RL-Guided Labeling}}
\label{alg:silver_rl_Guided}
\KwIn{Shapley vectors $(\Phi_{s_1}, \Phi_{s_2}, \dots, \Phi_{s_m})$, corresponding states $(s_1, s_2, \dots, s_m)$, trained RL policy $\pi$}
\KwParam{Number of discrete actions $k$}
\KwOut{Interpretable decision boundary functions $\{f_{ij}\}$ for each pair of actions $(i,j)$}

\textbf{Step 1: Action Clustering}\\
$A = \{A_1, \dots, A_k\} \leftarrow \text{K-Means}(\{\Phi_{s_i}\}_{i=1}^m, k)$\;
\For{$i = 1$ \KwTo $k$}{
    $\mu_i \leftarrow \frac{1}{|A_i|} \sum_{\Phi \in A_i} \Phi$ \tcp*[r]{Compute cluster centroids}
}

\textbf{Step 2: Boundary Points Identification}\\
Initialize empty set of boundary points $B \leftarrow \{\}$\;
\For{$i = 1$ \KwTo $k-1$}{
    \For{$j = i+1$ \KwTo $k$}{
        $X_{ij} \leftarrow \arg\min_{X \in \{\Phi_s\}} \big|\|X - \mu_i\|^2 - \|X - \mu_j\|^2\big|$ \tcp*[r]{Find geometric boundary point}
        $s_{ij} \leftarrow \phi^{-1}(X_{ij})$ \tcp*[r]{Inverse mapping to state space}
        $a_{ij} \leftarrow \pi(s_{ij})$ \tcp*[r]{Query RL policy for action label}
        $B \leftarrow B \cup \{(s_{ij}, a_{ij})\}$ \tcp*[r]{Store labeled boundary point}
    }
}

\textbf{Step 3: Interpretable Models Fitting}\\
$f_{ij}(s) \leftarrow \text{Fit Interpretable Model}(B)$\;

\textbf{Step 4: Output}\\
\Return{$\{f_{ij}\}$ as pairwise decision boundary functions}\;
\end{algorithm}

\clearpage

\section{Appendix B: Comparison between SILVER and SILVER with RL-Guided Labeling} \label{Appendix B: Comparison between SILVER and SILVER with RL-Guided Labeling}

\begin{table}[h]
\centering
\renewcommand{\arraystretch}{1.2}
\small
\begin{tabularx}{\textwidth}{p{3.3cm} X X}
\toprule
\textbf{Aspect} & \textbf{SILVER~\cite{li2025explainability}} & \textbf{SILVER with RL-Guided Labeling} \\
\midrule
\textbf{Target environments} & Restricted to binary-action, low-dimensional tasks (e.g., CartPole, MountainCar) & Scalable to high-dimensional, multi-action environments (up to 18 discrete actions) \\
\textbf{Boundary identification} & Determined purely by geometric distances between action clusters in Shapley vectors & Retains geometric identification but assigns action labels to each boundary state based on the output of RL policy $\pi(s_{ij})$ \\
\textbf{Labeling mechanism} & Implicitly determined by nearest cluster pairs in Shapley vectors & Explicitly determined by querying the RL policy for action labels \\
\textbf{Resulting data} & Boundary samples may be ambiguous or misaligned with the actual RL policy behavior & Behaviorally consistent dataset $(s_{ij}, a_{ij})$ enabling reliable interpretable models training  \\
\bottomrule
\end{tabularx}
\caption{Comparison between SILVER and SILVER with RL-Guided Labeling}
\label{tab:silver_vs_rlGuided}
\end{table}

\clearpage

\section{Appendix C: Gameplay Clips and Feature Extractor} \label{Appendix C: Game Clips and Feature Extractor}

\begin{figure}[!h]
    \centering
    \includegraphics[width=0.7\linewidth]{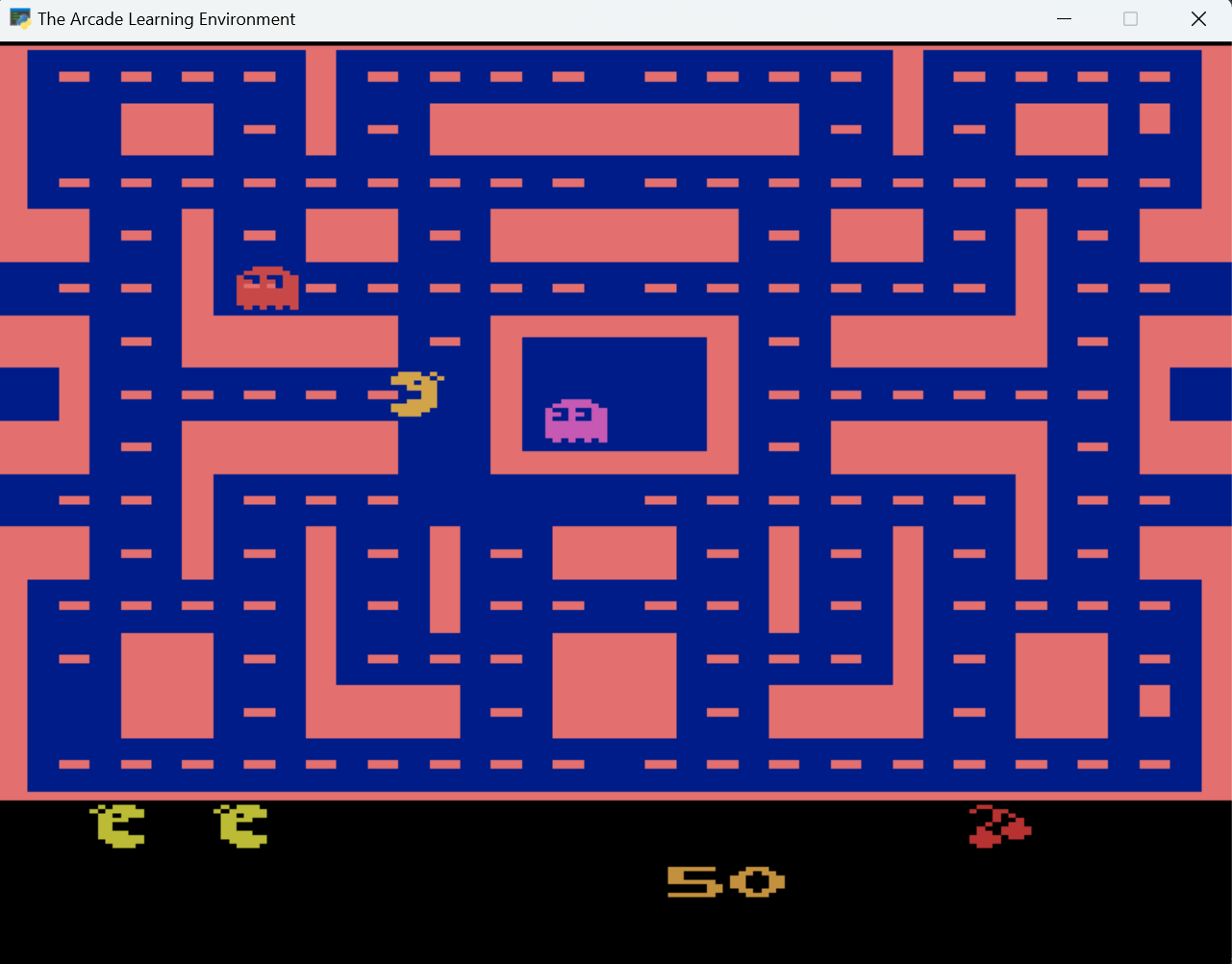}
    \caption{Gameplay clips of PPO in MsPacmanNoFrameskip-v4}
    \label{fig:MsPacman_gameclips}
\end{figure}

\begin{figure}[!h]
    \centering
    \includegraphics[width=0.7\linewidth]{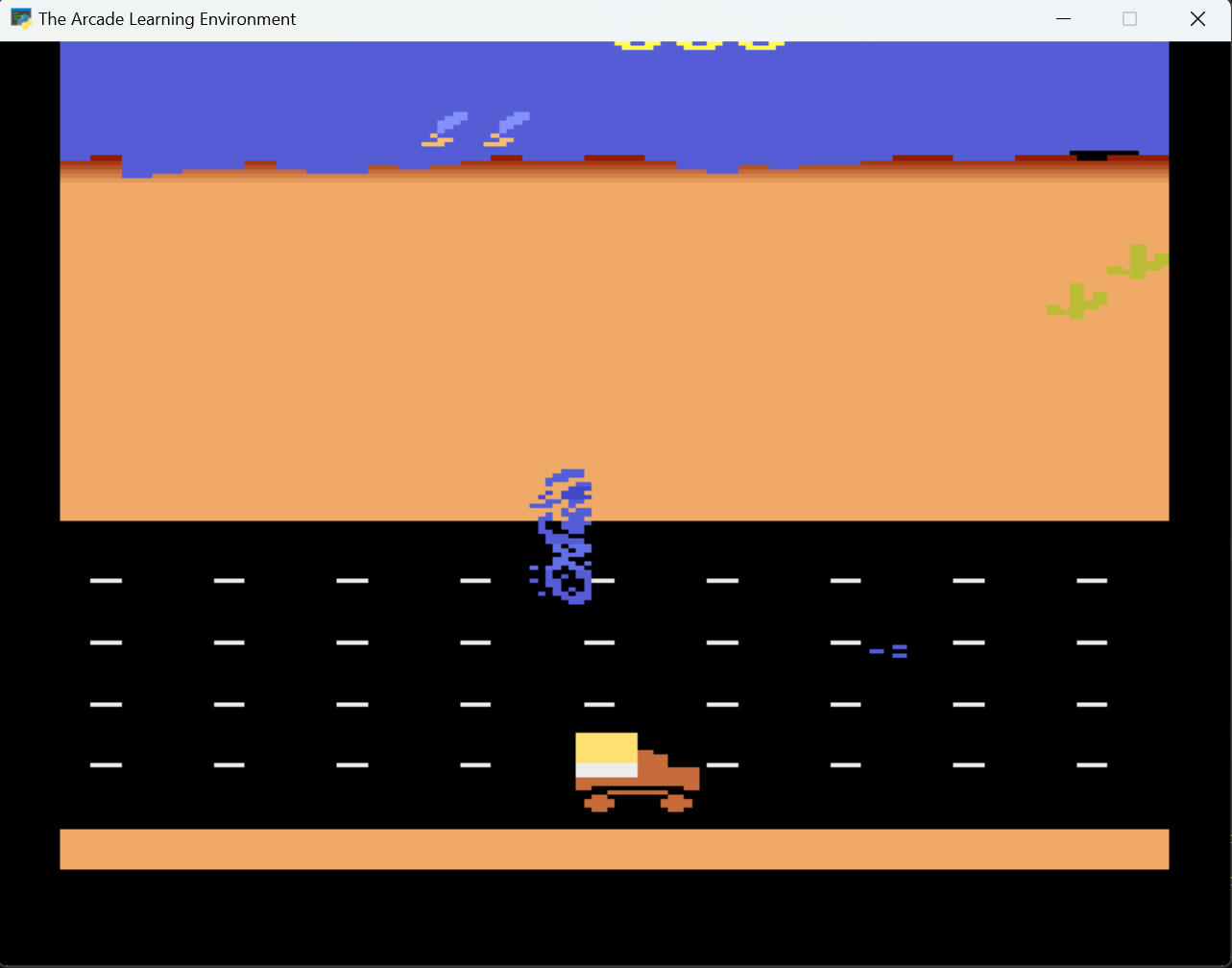}
    \caption{Gameplay clips of PPO in RoadRunnerNoFrameskip-v4 }
    \label{fig:RoadRunner_gameclips}
\end{figure}

\begin{figure}[ht]
    \centering
    \includegraphics[width=2.2\linewidth]{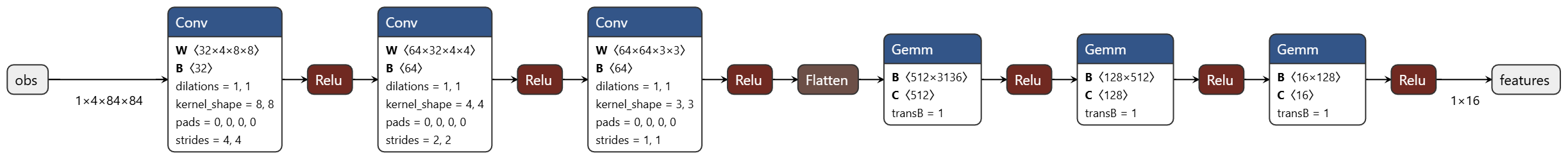}
    \caption{CNN-based feature extractor}
    \label{fig: Features Extractor}
\end{figure}

\clearpage

\section{Appendix D: Experiment on RoadRunner} \label{Appendix D: Experiment on RoadRunner}

The RoadRunner environment is a fast-paced Atari game where the agent controls the roadrunner to evade the pursuing coyote across desert terrain. The observation space consists of high-dimensional RGB frames depicting the roadrunner, the coyote, bird seed, and obstacles. The action space includes 18 discrete controls for movement, jumping, ducking, and firing. The agent gains rewards for collecting bird seed and avoiding capture, and penalties for collisions or being caught. The objective is to maximize survival while efficiently collecting rewards.

The contributions of state features to RL policy’s decisions are visualized in Figure~\ref{fig:RoadRunner_SHAP_comparison}. Based on these Shapley representations, we applied k-means clustering to the 18-action space and identified 153 boundary points that separate distinct behavioral regions.

Tables \ref{tab:RoadRunner_DQN_boundary_points}, \ref{tab:RoadRunner_DQN_shap_inverse_boundary_points}, \ref{tab:RoadRunner_A2C_shap_inverse_boundary_points}, \ref{tab:RoadRunner_A2C_boundary_points}, \ref{tab:RoadRunner_A2C_shap_inverse_boundary_points}, \ref{tab:RoadRunner_PPO_boundary_points}, \ref{tab:RoadRunner_PPO_shap_inverse_boundary_points} report the boundary points and their Shapley-inverse counterparts for each deep RL algorithm.

\begin{figure}[htbp]
    \centering
    \includegraphics[width=\textwidth]{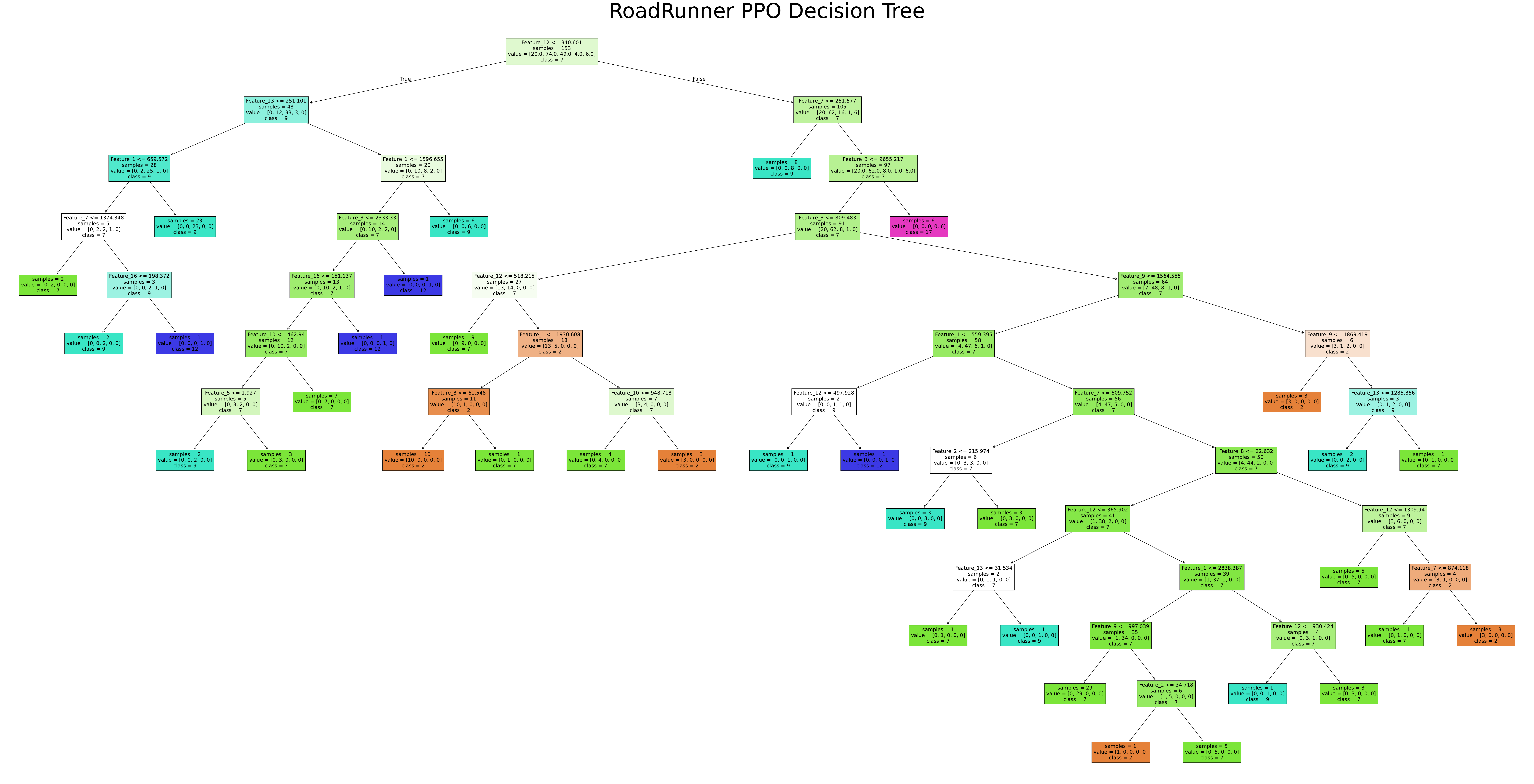}
    \caption{RoadRunner PPO decision tree}
    \label{fig:RoadRunner_ppo_decision_tree}
\end{figure}

\begin{figure}[htbp]
    \centering
    \includegraphics[width=\textwidth]{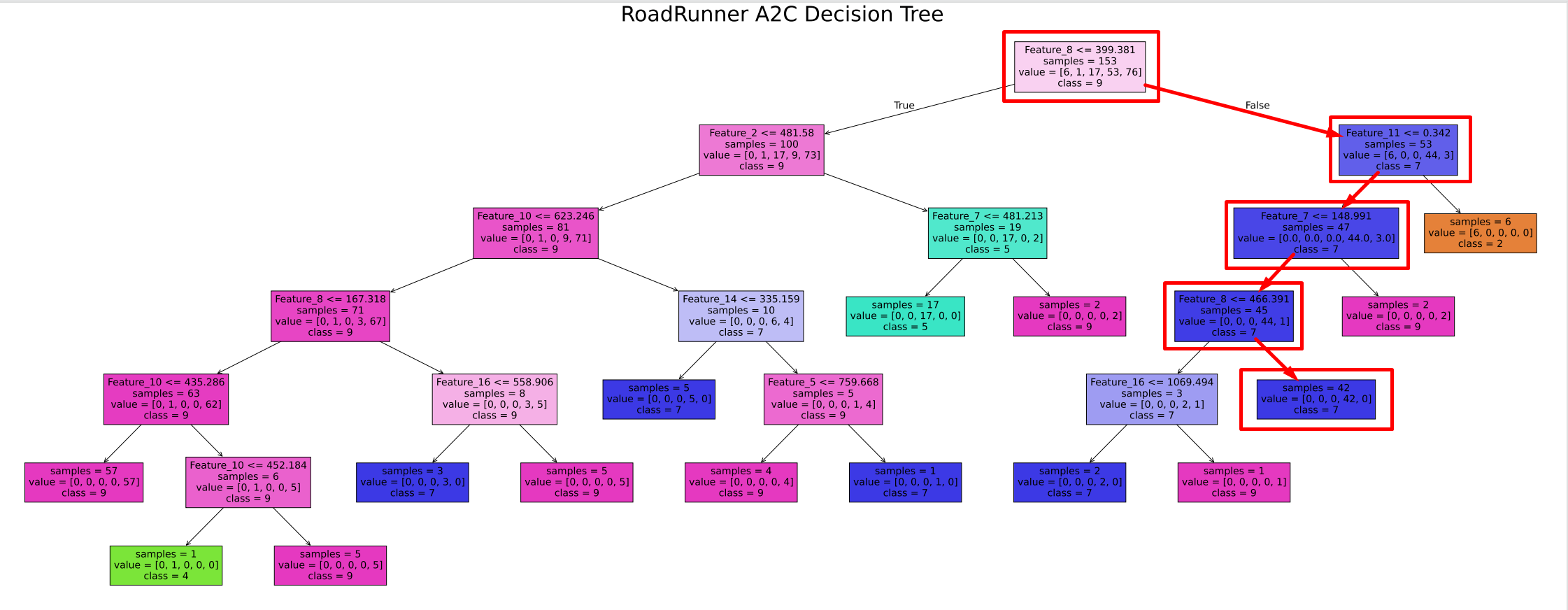}
    \caption{RoadRunner A2C decision tree}
    \label{fig:RoadRunner_a2c_decision_tree}
\end{figure}

\begin{figure}[htbp]
    \centering
    \includegraphics[width=\textwidth]{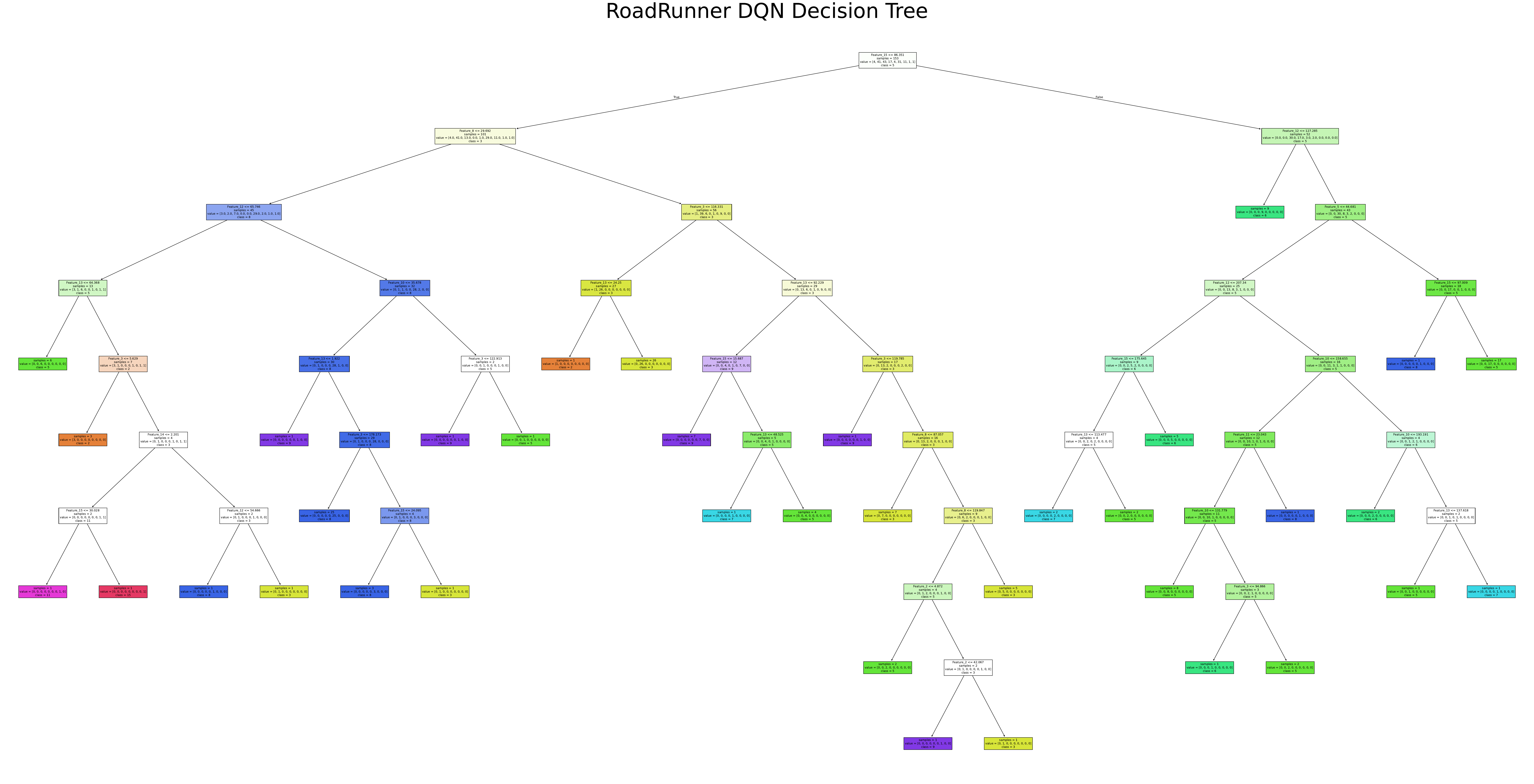}
    \caption{RoadRunner DQN decision tree}
    \label{fig:RoadRunner_dqn_decision_tree}
\end{figure}

\paragraph{Decision Tree} 
Figure~\ref{fig:RoadRunner_ppo_decision_tree}, \ref{fig:RoadRunner_a2c_decision_tree}, and \ref{fig:RoadRunner_dqn_decision_tree} present decision trees for PPO, A2C, and DQN, respectively. 

Taking A2C as an example, consider the first row in Table~\ref{tab:RoadRunner_A2C_shap_inverse_boundary_points}. Starting from the root node of Figure~\ref{fig:RoadRunner_a2c_decision_tree}, the decision process unfolds as follows:

\begin{enumerate}
  \item At the root node, check whether $Feature_{8} \leq 399.381$.\\
        Since $1037.385 > 399.381$, the process moves to the \textbf{right child}.
  \item Next, check whether $Feature_{11} \leq 0.342$.\\
        Since $0 < 0.342$, the process moves to the \textbf{left child}.
  \item Then, check whether $Feature_{7} \leq 148.991$.\\
        Since $0 < 148.991$, the process moves to the \textbf{left child}.
  \item Finally, check whether $Feature_{8} \leq 466.391$.\\
        Since $1037.385 > 466.391$, the process moves to the \textbf{right child}, reach a leaf: 
        $\boxed{\text{class} = 7}$, which corresponds to  $\boxed{\text{action} =  7}$.
\end{enumerate}

\paragraph{Linear Regression}
Table~\ref{tab:RoadRunner Linear Regression} illustrates the interpretable policy boundaries derived from linear regression for PPO, A2C, and DQN. Here, $x_{1}$ through $x_{16}$ represent the 16 state features used in the analysis.

Taking A2C as an example, consider the second row in Table~\ref{tab:RoadRunner_A2C_shap_inverse_boundary_points}. Substituting the feature values into the decision boundary in Table~\ref{tab:RoadRunner Linear Regression}, we obtain:

\[
\begin{aligned}
f(x) &= 8.82 \\
&\quad - (9.00 \times 10^{-4}) \times 95.19944 \\
&\quad - (6.94 \times 10^{-4}) \times 470.2562 \\
&\quad - (1.03 \times 10^{-3}) \times 180.0636 \\
&\quad - (4.19 \times 10^{-4}) \times 88.72929 \\
&\quad + (1.34 \times 10^{-3}) \times 96.3101 \\
&\quad + (1.08 \times 10^{-3}) \times 647.2335 \\
&\approx 9.01.
\end{aligned}
\]

Taking the floor of this value yields $\boxed{f(x) = 9}$, which corresponds to $\boxed{\text{action} =  9}$.

\paragraph{Logistic Regression}
Table~\ref{tab:RoadRunne_A2C_LogReg} illustrates the interpretable policy boundaries derived from logistic regression for A2C. 
Taking the third row of features in Table~\ref{tab:RoadRunner_A2C_shap_inverse_boundary_points}, we substitute the values into the corresponding decision boundaries. 
This yields a set of logits $\{f_{2}(x), f_{4}(x), f_{5}(x), f_{7}(x), f_{9}(x)\}$, each associated with one action. The procedure is identical to that already demonstrated for the MsPacman environment.

\clearpage

\section{Appendix E: Boundary Points and SHAP Inverse Boundary Points} \label{Appendix E: Boundary Points and SHAP Inverse Boundary Points}

\begin{table}[!h]
\centering
\scriptsize
\begin{minipage}{\textwidth}
\centering
\begin{tabular}{|rrrrrrrr|}
\hline
Feature 1 & Feature 2 & Feature 3 & Feature 4 & Feature 5 & Feature 6 & Feature 7 & Feature 8 \\
\hline
0.053718 & -0.58133 & -1.57969 & -10.0231 & 10.97137 & -3.29043 & -0.34815 & -0.07359 \\
-0.13582 & 0.166289 & 17.79752 & -8.14196 & 16.23918 & 0.061392 & 0.271136 & 362.5037 \\
-0.06903 & -0.93946 & 21.95477 & 194.3563 & 54.4827 & -3.42446 & -0.39898 & 104.8359 \\
0.147134 & -0.61962 & 21.12769 & 11.61575 & 41.8084 & -5.48569 & -0.59579 & 61.4086 \\
-0.08426 & -0.2127 & -3.38793 & -5.58269 & -26.1117 & -0.91717 & -0.01381 & -45.8628 \\
-0.09114 & -0.1284 & 18.9745 & -8.89167 & 29.9608 & -2.01041 & -0.14159 & 269.8545 \\
0.108812 & -0.73049 & 6.476514 & -10.0114 & 31.2885 & -3.58155 & -0.38707 & -6.30191 \\
0.250599 & -1.08509 & 16.77061 & 79.48419 & 36.1719 & -5.36885 & -0.59214 & 69.85063 \\
-0.10688 & -0.31807 & 14.7685 & -10.9438 & 26.3507 & -3.13208 & -0.13384 & 222.8886 \\
-0.03265 & -0.36164 & 1.063268 & -10.739 & 17.4609 & -2.72974 & -0.17212 & 68.40194 \\
\hline
\end{tabular}

\vspace{0.5cm}

\begin{tabular}{|rrrrrrrrr|}
\hline
Feature 9 & Feature 10 & Feature 11 & Feature 12 & Feature 13 & Feature 14 & Feature 15 & Feature 16 & Action Label \\
\hline
-28.4395 & -9.43087 & -0.56828 & 0.880025 & 93.17239 & -66.1944 & -1.10136 & 1.182046 & 6 \\
-28.2911 & -8.78055 & -28.6062 & 79.80954 & -71.0975 & -74.9427 & -5.54564 & 1.781113 & 2 \\
131.3282 & 0.75164 & 85.30825 & 0.966173 & 48.69307 & 179.52 & -1.15084 & 2.031596 & 7 \\
46.60157 & 47.24052 & 92.89508 & -26.421 & 183.818 & 43.83136 & -2.52363 & 1.589851 & 6 \\
-13.59 & -5.24955 & 24.39548 & -0.68877 & -17.2954 & -45.0942 & -0.00598 & 0.627361 & 6 \\
-17.6029 & 13.41057 & 10.66782 & 36.58551 & -82.5216 & 31.31346 & -2.97744 & 1.89597 & 7 \\
-29.8968 & 13.4846 & 26.46422 & 1.507363 & 112.6223 & 10.08722 & -1.94405 & 1.693566 & 6 \\
-18.2203 & 92.6196 & 153.8688 & 0.966203 & 249.4967 & -31.6483 & -1.70886 & 1.296677 & 7 \\
-36.1316 & 15.26726 & 13.5926 & -2.39442 & -98.3716 & 53.67466 & -1.89186 & 2.008542 & 7 \\
-32.6094 & -7.19312 & -23.782 & -2.90944 & -94.4912 & -0.48619 & -2.44242 & 1.7049 & 5 \\
\hline
\end{tabular}
\caption{MsPacman DQN boundary points: first 10 rows of features and action labels}
\label{tab:MsPacman_DQN_boundary_points}
\end{minipage}
\end{table}

\begin{table*}[!h]
\centering
\scriptsize
\begin{tabular}{|rrrrrrrr|}
\hline
Feature 1 & Feature 2 & Feature 3 & Feature 4 & Feature 5 & Feature 6 & Feature 7 & Feature 8 \\
\hline
0 & 0 & 0 & 0 & 55.7541 & 0 & 0 & 721.5164 \\
0 & 0 & 0 & 0 & 0 & 0 & 0 & 55.56659 \\
0 & 0 & 0 & 501.7624 & 0 & 0 & 0 & 336.9333 \\
0 & 0 & 0 & 85.61454 & 0 & 0 & 0 & 431.6569 \\
0 & 0 & 0 & 0 & 489.7511 & 0 & 0 & 1842.243 \\
0 & 0 & 0 & 0 & 0 & 0 & 0 & 137.5602 \\
0 & 0 & 0 & 0 & 0 & 0 & 0 & 830.2186 \\
0 & 0 & 0 & 255.8091 & 0 & 0 & 0 & 336.7449 \\
0 & 0 & 0 & 0 & 0 & 0 & 0 & 134.5811 \\
0 & 0 & 0 & 0 & 0 & 0 & 0 & 321.8050 \\
\hline
\end{tabular}

\vspace{0.5cm}

\begin{tabular}{|rrrrrrrrr|}
\hline
Feature 9 & Feature 10 & Feature 11 & Feature 12 & Feature 13 & Feature 14 & Feature 15 & Feature 16 & Action Label \\
\hline
0 & 0 & 318.1658 & 0 & 545.9904 & 84.66984 & 0 & 0 & 6 \\
0 & 11.0026 & 0 & 304.9976 & 0 & 122.1123 & 0 & 0 & 2 \\
300.3364 & 18.30009 & 207.49 & 0 & 457.2445 & 438.9895 & 0 & 0 & 7 \\
180.606 & 132.7699 & 86.33698 & 81.04003 & 638.9478 & 276.7383 & 0 & 0 & 6 \\
0 & 0 & 1569.231 & 0 & 275.1934 & 0 & 0 & 0 & 6 \\
50.81373 & 59.32636 & 0 & 270.8656 & 28.70813 & 270.455 & 0 & 0 & 7 \\
0 & 0 & 229.8056 & 0 & 536.8198 & 260.828 & 0 & 0 & 6 \\
140.1302 & 42.31081 & 58.60357 & 0 & 670.4374 & 176.3808 & 0 & 0 & 7 \\
0 & 30.09257 & 104.6578 & 0 & 0 & 299.6935 & 0 & 0 & 7 \\
0 & 0 & 262.2588 & 0 & 50.81428 & 240.3015 & 0 & 0 & 5 \\
\hline
\end{tabular}
\caption{MsPacman DQN SHAP inverse boundary points: first 10 rows of features and action labels}
\label{tab:MsPacman_DQN_shap_inverse_boundary_points}
\end{table*}

\begin{table*}[!h]
\centering
\scriptsize
\begin{tabular}{|rrrrrrrr|}
\hline
Feature 1 & Feature 2 & Feature 3 & Feature 4 & Feature 5 & Feature 6 & Feature 7 & Feature 8 \\
\hline
0 & -0.00083 & -0.47892 & 1.12E-05 & 3.01E-05 & 0 & 0 & 0.058596 \\
0 & 0.000949 & -0.27889 & -1.53E-05 & 1.98E-05 & 0 & 0 & -0.1481 \\
0 & -0.00135 & 0.477397 & -5.52E-08 & 6.72E-06 & 0 & 0 & 0.135079 \\
0 & -0.00022 & 0.081397 & -2.44E-06 & -1.36E-05 & 0 & 0 & -0.29276 \\
0 & -0.00043 & -0.24212 & 1.18E-05 & 7.31E-05 & 0 & 0 & -0.29675 \\
0 & 0.000124 & 2.800792 & -2.25E-05 & -8.00E-05 & 0 & 0 & 0.189882 \\
0 & -0.00087 & -0.57139 & 1.17E-05 & 2.97E-05 & 0 & 0 & 0.039338 \\
0 & 9.14E-05 & 3.346151 & -3.30E-05 & 5.00E-06 & 0 & 0 & 0.127679 \\
0 & -0.0001 & 1.135439 & -2.20E-05 & 3.00E-05 & 0 & 0 & -0.50111 \\
0 & -0.00041 & -0.2112 & 4.40E-06 & 1.48E-05 & 0 & 0 & 0.028318 \\
\hline
\end{tabular}

\vspace{0.5cm}

\begin{tabular}{|rrrrrrrrr|}
\hline
Feature 9 & Feature 10 & Feature 11 & Feature 12 & Feature 13 & Feature 14 & Feature 15 & Feature 16 & Action Label \\
\hline
-0.00458 & -0.28035 & 0.176815 & -1.37E-05 & 0 & -0.20625 & 0 & 0 & 5 \\
0.179629 & -0.16501 & 2.131016 & 0.000116 & 0 & 0.10562 & 0 & 0 & 7 \\
-0.00406 & 0.423494 & -1.38735 & -1.44E-05 & 0 & -0.00709 & 0 & 0 & 5 \\
-0.0016 & -0.46054 & -0.36306 & -4.85E-05 & 0 & 0.168933 & 0 & 0 & 5 \\
-0.00646 & 0.14528 & 1.490529 & 6.42E-05 & 0 & 0.148152 & 0 & 0 & 5 \\
0.001417 & -0.01967 & 1.789025 & 8.46E-05 & 0 & 0.37831 & 0 & 0 & 8 \\
-0.00475 & -0.34087 & 0.33842 & -1.48E-05 & 0 & -0.19539 & 0 & 0 & 5 \\
0.001948 & -0.01433 & 1.796733 & 0.00011 & 0 & 0.671487 & 0 & 0 & 8 \\
0.000805 & -0.07721 & 3.341121 & 7.21E-05 & 0 & 0.606315 & 0 & 0 & 8 \\
-0.0022 & -0.01824 & -0.36886 & -1.40E-05 & 0 & -0.16291 & 0 & 0 & 5 \\
\hline
\end{tabular}
\caption{MsPacman A2C boundary points: first 10 rows of features and action labels}
\label{tab:MsPacman_A2C_boundary_points}
\end{table*}

\begin{table*}[!h]
\centering
\scriptsize
\begin{tabular}{|rrrrrrrr|}
\hline
Feature 1 & Feature 2 & Feature 3 & Feature 4 & Feature 5 & Feature 6 & Feature 7 & Feature 8 \\
\hline
0 & 0 & 623.6633 & 0 & 0 & 0 & 0 & 532.5274 \\
0 & 0 & 653.3041 & 0 & 0 & 0 & 0 & 107.4317 \\
0 & 0 & 1341.658 & 0 & 0 & 0 & 0 & 727.1529 \\
0 & 0 & 1061.923 & 0 & 0 & 0 & 0 & 494.4895 \\
0 & 0 & 616.0728 & 0 & 0 & 0 & 0 & 259.9639 \\
0 & 0 & 478.1401 & 0 & 0 & 0 & 0 & 696.2931 \\
0 & 0 & 583.0953 & 0 & 0 & 0 & 0 & 501.0357 \\
0 & 0 & 384.8195 & 0 & 0 & 0 & 0 & 688.3864 \\
0 & 0 & 710.8715 & 0 & 0 & 0 & 0 & 983.8112 \\
0 & 0 & 568.2379 & 0 & 0 & 0 & 0 & 675.8421 \\
\hline
\end{tabular}

\vspace{0.5cm}

\begin{tabular}{|rrrrrrrrr|}
\hline
Feature 9 & Feature 10 & Feature 11 & Feature 12 & Feature 13 & Feature 14 & Feature 15 & Feature 16 & Action Label \\
\hline
0 & 150.9219 & 601.0638 & 0 & 0 & 162.5402 & 0 & 0 & 5 \\
87.21083 & 81.74716 & 0 & 0 & 0 & 0 & 0 & 0 & 7 \\
0 & 668.9421 & 1414.396 & 0 & 0 & 64.81534 & 0 & 0 & 5 \\
0 & 93.9263 & 1125.77 & 0 & 0 & 0 & 0 & 0 & 5 \\
0 & 230.8709 & 124.3961 & 0 & 0 & 0 & 0 & 0 & 5 \\
0 & 274.8654 & 1496.319 & 0 & 0 & 441.1012 & 0 & 0 & 8 \\
0 & 121.8064 & 550.3 & 0 & 0 & 147.2741 & 0 & 0 & 5 \\
0 & 166.9695 & 1494.9 & 0 & 0 & 517.485 & 0 & 0 & 8 \\
0 & 330.6453 & 2159.496 & 0 & 0 & 576.4459 & 0 & 0 & 8 \\
0 & 131.7175 & 958.8807 & 0 & 0 & 385.6672 & 0 & 0 & 5 \\
\hline
\end{tabular}
\caption{MsPacman A2C SHAP inverse boundary points: first 10 rows of features and action labels}
\label{tab:MsPacman_A2C_shap_inverse_boundary_points}
\end{table*}

\begin{table*}[!h]
\centering
\scriptsize
\begin{tabular}{|rrrrrrrr|}
\hline
Feature 1 & Feature 2 & Feature 3 & Feature 4 & Feature 5 & Feature 6 & Feature 7 & Feature 8 \\
\hline
0.000 & -0.28516 & 0.000 & 0.072798 & 0.000265 & 0.000 & 0.000 & 0.000 \\
0.000 & -0.08226 & 0.000 & 0.378046 & -0.00056 & 0.000 & 0.000 & 0.000 \\
0.000 & 0.86831 & 0.000 & 0.295693 & -0.00029 & 0.000 & 0.000 & 0.000 \\
0.000 & -0.58381 & 0.000 & 0.084519 & -0.000063 & 0.000 & 0.000 & 0.000 \\
0.000 & -0.37113 & 0.000 & -0.08261 & -0.00043 & 0.000 & 0.000 & 0.000 \\
0.000 & -0.23301 & 0.000 & -0.11092 & -0.00054 & 0.000 & 0.000 & 0.000 \\
0.000 & -0.3258 & 0.000 & 0.177798 & 0.00016 & 0.000 & 0.000 & 0.000 \\
0.000 & -0.13854 & 0.000 & -0.07107 & -0.00072 & 0.000 & 0.000 & 0.000 \\
0.000 & -0.39994 & 0.000 & -0.07986 & -0.00044 & 0.000 & 0.000 & 0.000 \\
0.000 & 0.147544 & 0.000 & -0.07974 & -0.00096 & 0.000 & 0.000 & 0.000 \\
\hline
\end{tabular}

\vspace{0.5cm}

\begin{tabular}{|rrrrrrrrr|}
\hline
Feature 9 & Feature 10 & Feature 11 & Feature 12 & Feature 13 & Feature 14 & Feature 15 & Feature 16 & Action Label \\
\hline
0.000 & -0.4569 & -0.01058 & 2.505064 & 0.000 & 0.000 & 0.26214 & 0.000 & 7 \\
0.000 & 0.34163 & 0.010641 & 1.282444 & 0.000 & 0.000 & -0.04949 & 0.000 & 6 \\
0.000 & 0.68037 & 0.010819 & 0.913588 & 0.000 & 0.000 & -0.01256 & 0.000 & 6 \\
0.000 & -0.61006 & -0.0000955 & 3.774243 & 0.000 & 0.000 & -0.01605 & 0.000 & 7 \\
0.000 & -0.30679 & 0.004481 & 1.660903 & 0.000 & 0.000 & 0.040685 & 0.000 & 6 \\
0.000 & -0.33712 & 0.004529 & 1.43017 & 0.000 & 0.000 & 0.06036 & 0.000 & 6 \\
0.000 & -0.43532 & -0.00619 & 2.333613 & 0.000 & 0.000 & -0.03847 & 0.000 & 7 \\
0.000 & -0.49607 & 0.014154 & 2.157 & 0.000 & 0.000 & 0.012373 & 0.000 & 6 \\
0.000 & -0.64253 & 0.004522 & 1.725279 & 0.000 & 0.000 & 0.238071 & 0.000 & 6 \\
0.000 & -0.35632 & 0.015381 & 1.709701 & 0.000 & 0.000 & 0.350303 & 0.000 & 6 \\
\hline
\end{tabular}
\caption{MsPacman PPO boundary points: first 10 rows of features and action labels}
\label{tab:MsPacman_PPO_boundary_points}
\end{table*}

\begin{table*}[!h]
\centering
\scriptsize
\begin{tabular}{|rrrrrrrr|}
\hline
Feature 1 & Feature 2 & Feature 3 & Feature 4 & Feature 5 & Feature 6 & Feature 7 & Feature 8 \\
\hline
0.0 & 5753.206 & 0.0 & 530.9578 & 0.0 & 0.0 & 0.0 & 0.0 \\
0.0 & 6628.804 & 0.0 & 1914.645 & 0.0 & 0.0 & 0.0 & 0.0 \\
0.0 & 0.000 & 0.0 & 2528.653 & 0.0 & 0.0 & 0.0 & 0.0 \\
0.0 & 4611.498 & 0.0 & 550.1152 & 0.0 & 0.0 & 0.0 & 0.0 \\
0.0 & 8737.567 & 0.0 & 0.000 & 0.0 & 0.0 & 0.0 & 0.0 \\
0.0 & 7480.565 & 0.0 & 0.000 & 0.0 & 0.0 & 0.0 & 0.0 \\
0.0 & 5501.599 & 0.0 & 33.12456 & 0.0 & 0.0 & 0.0 & 0.0 \\
0.0 & 7509.492 & 0.0 & 0.000 & 0.0 & 0.0 & 0.0 & 0.0 \\
0.0 & 8891.553 & 0.0 & 0.000 & 0.0 & 0.0 & 0.0 & 0.0 \\
0.0 & 6040.661 & 0.0 & 0.000 & 0.0 & 0.0 & 0.0 & 0.0 \\
\hline
\end{tabular}

\vspace{0.5cm}

\begin{tabular}{|rrrrrrrrr|}
\hline
Feature 9 & Feature 10 & Feature 11 & Feature 12 & Feature 13 & Feature 14 & Feature 15 & Feature 16 & Action Label \\
\hline
0.0 & 2318.034 & 0.0 & 8493.874 & 0.0 & 0.0 & 8382.263 & 0.0 & 7 \\
0.0 & 1880.883 & 0.0 & 14349.77 & 0.0 & 0.0 & 11766.16 & 0.0 & 6 \\
0.0 & 0.000 & 0.0 & 20786.43 & 0.0 & 0.0 & 9735.063 & 0.0 & 6 \\
0.0 & 1762.912 & 0.0 & 6748.198 & 0.0 & 0.0 & 6996.48 & 0.0 & 7 \\
0.0 & 6482.863 & 0.0 & 22678.59 & 0.0 & 0.0 & 3191.511 & 0.0 & 6 \\
0.0 & 4270.52 & 0.0 & 17331.35 & 0.0 & 0.0 & 6207.685 & 0.0 & 6 \\
0.0 & 2374.423 & 0.0 & 8500.413 & 0.0 & 0.0 & 7080.591 & 0.0 & 7 \\
0.0 & 6573.165 & 0.0 & 29362.53 & 0.0 & 0.0 & 1421.398 & 0.0 & 6 \\
0.0 & 6896.546 & 0.0 & 23498.77 & 0.0 & 0.0 & 3105.686 & 0.0 & 6 \\
0.0 & 5298.294 & 0.0 & 23890.88 & 0.0 & 0.0 & 1742.465 & 0.0 & 6 \\
\hline
\end{tabular}
\caption{MsPacman PPO SHAP inverse boundary points: first 10 rows of features and action labels}
\label{tab:MsPacman_PPO_shap_inverse_boundary_points}
\end{table*}

\begin{table*}[!h]
\centering
\scriptsize
\begin{tabular}{|rrrrrrrr|}
\hline
Feature 1 & Feature 2 & Feature 3 & Feature 4 & Feature 5 & Feature 6 & Feature 7 & Feature 8 \\
\hline
5.474527 & -0.81353 & 67.99976 & -0.18912 & -26.918 & -0.15756 & -1.73123 & -10.2803 \\
2.992853 & 2.733483 & -3.56914 & -0.13995 & 5.06779 & 0.207614 & -3.64042 & 25.05078 \\
1.108827 & -0.73966 & -10.4527 & -0.04402 & 8.2174 & -0.04362 & -2.02183 & -9.78841 \\
12.07903 & -0.79765 & 190.9837 & -0.37814 & 3.13685 & -0.28556 & -2.23423 & 106.7969 \\
10.21576 & 3.761232 & 68.81969 & -0.36309 & 4.5211 & -0.27273 & -3.14379 & 86.7971 \\
10.2965 & -0.78144 & 74.93028 & -0.32361 & 4.95804 & -0.19189 & -2.76325 & 86.90049 \\
8.507568 & 8.391128 & 151.9825 & -0.32951 & 12.85635 & -0.34459 & -4.24815 & -17.8449 \\
2.938708 & -0.72215 & 26.80925 & -0.13382 & 3.2263 & -0.06201 & -1.64585 & -11.6016 \\
15.82252 & -0.84811 & 209.365 & -0.43631 & 4.4694 & -0.31842 & -2.09036 & 125.2 \\
4.299641 & -0.94559 & -7.29464 & -0.18471 & 13.2794 & -0.28606 & 14.64754 & 19.93434 \\
\hline
\end{tabular}

\vspace{0.5cm}

\begin{tabular}{|rrrrrrrrr|}
\hline
Feature 9 & Feature 10 & Feature 11 & Feature 12 & Feature 13 & Feature 14 & Feature 15 & Feature 16 & Action Label \\
\hline
3.331623 & -9.77 & 8.657984 & 315.9 & 28.0846 & -1.42897 & 3.152705 & 0.0000242 & 8 \\
0.443141 & -5.35439 & 5.420255 & 0.539487 & -14.8831 & -0.15377 & -7.97344 & 0.0000534 & 3 \\
0.313884 & -6.69001 & 1.645254 & 44.13974 & -18.1456 & -0.80795 & -1.55171 & 0.0000535 & 8 \\
3.305505 & 13.59553 & 16.9441 & 8.958751 & -145.376 & 1.377216 & 34.02753 & 0.0000000 & 5 \\
2.244834 & 58.93856 & -1.35426 & -3.01145 & 4.090079 & -0.09288 & -9.17819 & 0.0000379 & 8 \\
1.037774 & 58.97981 & 13.57428 & 3.093952 & -80.4599 & 3.308704 & 12.90698 & 0.0000372 & 6 \\
3.216183 & -13.9531 & 15.05736 & 49.27177 & 89.38817 & -0.52471 & 11.5848 & 0.0000000 & 9 \\
1.637403 & -13.2263 & 5.023197 & 103.6954 & -21.3427 & 5.47379 & -8.87605 & 0.0000372 & 5 \\
3.872871 & -14.9694 & 20.44513 & 0.918966 & -64.6853 & 1.795213 & 53.85848 & 0.0000000 & 8 \\
-0.47081 & -12.7496 & 6.945717 & -7.97213 & 9.558705 & 3.580552 & -8.14629 & 0.000129 & 3 \\
\hline
\end{tabular}
\caption{RoadRunner DQN boundary points: first 10 rows of features and action labels}
\label{tab:RoadRunner_DQN_boundary_points}
\end{table*}

\begin{table*}[!h]
\centering
\scriptsize
\begin{tabular}{|rrrrrrrr|}
\hline
Feature 1 & Feature 2 & Feature 3 & Feature 4 & Feature 5 & Feature 6 & Feature 7 & Feature 8 \\
\hline
0.000 & 0.000 & 188.7657 & 0.000 & 208.6209 & 0.000 & 0.000 & 0.000 \\
0.000 & 13.94768 & 50.96701 & 0.000 & 18.80776 & 8.201801 & 0.000 & 60.99874 \\
0.000 & 0.000 & 35.64004 & 0.000 & 105.0581 & 0.000 & 0.000 & 0.000 \\
0.000 & 0.000 & 227.0123 & 0.000 & 95.6398 & 0.000 & 0.000 & 127.3856 \\
0.000 & 14.1228 & 113.9336 & 0.000 & 49.6401 & 0.000 & 0.000 & 106.4597 \\
0.000 & 0.000 & 138.5666 & 0.000 & 23.21588 & 0.000 & 0.000 & 110.8463 \\
0.000 & 27.12692 & 207.4527 & 0.000 & 0.000 & 0.000 & 0.000 & 0.000 \\
0.000 & 0.000 & 113.7223 & 0.000 & 128.9664 & 0.000 & 0.000 & 0.000 \\
0.000 & 0.000 & 231.0666 & 0.000 & 101.2544 & 0.000 & 0.000 & 138.9797 \\
0.000 & 0.000 & 42.31835 & 0.000 & 0.000 & 36.38669 & 0.000 & 53.29093 \\
\hline
\end{tabular}

\vspace{0.5cm}

\begin{tabular}{|rrrrrrrrr|}
\hline
Feature 9 & Feature 10 & Feature 11 & Feature 12 & Feature 13 & Feature 14 & Feature 15 & Feature 16 & Action Label \\
\hline
0.000 & 0.000 & 0.000 & 604.8348 & 57.50573 & 0.000 & 0.000 & 0.000 & 8 \\
0.000 & 21.39673 & 0.000 & 104.7776 & 110.2937 & 5.451507 & 0.000 & 0.000 & 3 \\
0.000 & 0.000 & 0.000 & 196.1525 & 121.0192 & 0.000 & 0.000 & 0.000 & 8 \\
0.000 & 186.7019 & 0.000 & 223.3806 & 273.6460 & 0.000 & 125.3206 & 0.000 & 5 \\
0.000 & 114.6822 & 7.669223 & 84.23914 & 96.00603 & 6.905884 & 0.000 & 0.000 & 3 \\
0.000 & 134.9267 & 0.000 & 204.6313 & 213.2327 & 0.000 & 221.8269 & 0.000 & 6 \\
0.000 & 7.785656 & 0.000 & 237.4772 & 0.000 & 3.684652 & 0.000 & 0.000 & 9 \\
0.000 & 0.000 & 0.000 & 285.1784 & 129.4608 & 53.5535 & 4.809501 & 0.000 & 8 \\
0.000 & 219.5916 & 0.000 & 165.9190 & 180.7253 & 0.000 & 168.3602 & 0.000 & 5 \\
0.000 & 0.000 & 0.000 & 81.0340 & 75.77461 & 0.000 & 9.164403 & 0.000 & 3 \\
\hline
\end{tabular}
\caption{RoadRunner DQN SHAP inverse boundary points: first 10 rows of features and action labels}
\label{tab:RoadRunner_DQN_shap_inverse_boundary_points}
\end{table*}

\begin{table*}[!h]
\centering
\scriptsize
\begin{tabular}{|rrrrrrrr|}
\hline
Feature 1 & Feature 2 & Feature 3 & Feature 4 & Feature 5 & Feature 6 & Feature 7 & Feature 8 \\
\hline
-0.0016 & 0.012969 & 0.002574 & 0.006918 & 0.090507 & 0.004858 & 0.344943 & 4.097872 \\
0.067005 & 0.111029 & -0.00269 & 0.056097 & 0.010497 & -0.0101 & 0.396967 & 2.144845 \\
0.168133 & 0.205574 & -0.00202 & 0.046697 & -0.085069 & -0.0076 & 0.334608 & 3.086863 \\
0.251399 & 0.11519 & -0.00046 & 0.061054 & -1.47655 & 0.007191 & 0.419466 & 2.611407 \\
0.172695 & 0.072814 & 0.001032 & 0.126251 & 2.13337 & 0.092112 & 0.115391 & 4.472525 \\
0.117376 & 0.144224 & -0.00195 & 0.032066 & -2.19087 & 0.00218 & 0.313983 & 2.340529 \\
0.057967 & 0.136672 & -0.00153 & -0.00688 & -2.59928 & 0.00963 & 0.348639 & 2.715359 \\
-0.00683 & 0.00922 & 0.001372 & -0.021748 & 0.149261 & 0.028857 & 0.319187 & 5.202232 \\
0.238766 & 1.495085 & 0.002403 & 0.017486 & 0.559244 & 0.006417 & 0.38527 & 1.023632 \\
-1.22581 & 0.046178 & -0.00249 & 0.054852 & 0.112022 & -0.0097 & -1.05545 & 1.078114 \\
\hline
\end{tabular}

\vspace{0.5cm}

\begin{tabular}{|rrrrrrrrr|}
\hline
Feature 9 & Feature 10 & Feature 11 & Feature 12 & Feature 13 & Feature 14 & Feature 15 & Feature 16 & Action Label \\
\hline
-0.00038 & 0.193474 & 0.009643 & 0.021532 & 0.188902 & 0.044162 & 0.022337 & 0.328965 & 7 \\
-0.00105 & 0.186545 & 0.011617 & 0.150055 & 0.042825 & -0.77204 & 0.028056 & 0.107762 & 9 \\
-0.00129 & 0.378642 & 0.009852 & 0.202114 & 0.154361 & -0.79497 & -0.02757 & -1.49946 & 9 \\
-0.00092 & 1.458449 & 0.021435 & 0.111257 & 0.220765 & -0.51995 & -0.00403 & -0.45147 & 9 \\
-0.00042 & 0.886021 & -0.02182 & 0.314617 & 0.63647 & 0.347686 & 0.0359 & -1.44858 & 2 \\
-0.00008 & 0.033347 & 0.005236 & 0.12489 & 0.057393 & -0.38435 & -0.02944 & -0.82626 & 9 \\
-0.00013 & 0.112104 & 0.008091 & 0.155574 & 0.145118 & -0.62689 & -0.04284 & -1.12652 & 9 \\
-0.00018 & 0.272794 & 0.007354 & -0.03508 & 0.13459 & -0.02189 & -0.01719 & -0.19001 & 7 \\
0.015837 & 0.64423 & 0.001668 & -0.62335 & 2.441334 & -1.68436 & 0.08042 & 0.177791 & 5 \\
-0.0002 & 0.139825 & -0.000057 & -3.04481 & 0.041828 & -1.05401 & -0.00218 & 6.010482 & 9 \\
\hline
\end{tabular}
\caption{RoadRunner A2C boundary points: first 10 rows of features and action labels}
\label{tab:RoadRunner_A2C_boundary_points}
\end{table*}

\begin{table*}[!h]
\centering
\scriptsize
\begin{tabular}{|rrrrrrrr|}
\hline
Feature 1 & Feature 2 & Feature 3 & Feature 4 & Feature 5 & Feature 6 & Feature 7 & Feature 8 \\
\hline
161.505 & 0.000 & 0.000 & 0.000 & 849.9095 & 0.000 & 0.000 & 1037.385 \\
95.19944 & 0.000 & 0.000 & 0.000 & 470.2562 & 0.000 & 0.000 & 0.000 \\
66.96087 & 0.000 & 0.000 & 0.000 & 172.4873 & 0.000 & 6.232745 & 0.000 \\
51.1582 & 50.33921 & 0.000 & 0.000 & 62.84172 & 237.3442 & 33.74626 & 936.9337 \\
549.525 & 101.9358 & 0.000 & 0.000 & 104.4528 & 0.000 & 0.000 & 0.000 \\
89.66468 & 0.000 & 0.000 & 0.000 & 93.95311 & 0.000 & 35.43744 & 0.000 \\
131.0036 & 0.000 & 0.000 & 0.000 & 7.160192 & 0.000 & 54.5494 & 0.000 \\
90.4573 & 0.000 & 0.000 & 0.000 & 132.6835 & 187.1228 & 0.000 & 2613.281 \\
13.69985 & 847.7397 & 0.000 & 0.000 & 468.4573 & 19.38743 & 0.000 & 0.000 \\
616.4103 & 0.000 & 0.000 & 0.000 & 180.913 & 0.000 & 645.9976 & 0.000 \\
\hline
\end{tabular}

\vspace{0.5cm}

\begin{tabular}{|rrrrrrrrr|}
\hline
Feature 9 & Feature 10 & Feature 11 & Feature 12 & Feature 13 & Feature 14 & Feature 15 & Feature 16 & Action Label \\
\hline
0.000 & 440.8918 & 0.000 & 0.000 & 0.000 & 270.6354 & 56.48066 & 786.6356 & 7 \\
0.000 & 180.0636 & 0.000 & 0.000 & 0.000 & 88.72929 & 96.3101 & 647.2335 & 9 \\
0.000 & 171.6217 & 0.000 & 0.000 & 0.000 & 110.6941 & 0.000 & 181.6488 & 9 \\
0.000 & 74.64204 & 55.82757 & 964.7496 & 246.0649 & 106.0566 & 0.000 & 317.4086 & 9 \\
0.000 & 247.9345 & 0.000 & 0.000 & 0.000 & 157.3781 & 21.54585 & 1578.259 & 2 \\
0.000 & 145.2424 & 0.000 & 0.000 & 0.000 & 45.90677 & 0.000 & 229.8769 & 9 \\
0.000 & 486.9624 & 0.000 & 502.4604 & 0.000 & 159.6311 & 460.3797 & 1833.193 & 7 \\
0.000 & 0.000 & 0.000 & 332.5967 & 611.1736 & 818.4494 & 531.9953 & 2331.473 & 5 \\
0.000 & 0.000 & 0.000 & 1056.952 & 0.000 & 3.037189 & 0.000 & 3703.815 & 9 \\
\hline
\end{tabular}
\caption{RoadRunner A2C SHAP inverse boundary points: first 10 rows of features and action labels}
\label{tab:RoadRunner_A2C_shap_inverse_boundary_points}
\end{table*}

\begin{table*}[!h]
\centering
\scriptsize
\begin{tabular}{|rrrrrrrr|}
\hline
Feature 1 & Feature 2 & Feature 3 & Feature 4 & Feature 5 & Feature 6 & Feature 7 & Feature 8 \\
\hline
-4.60653 & 0.296192 & 0.969315 & 0.000136 & 0.08256 & 0.000 & 2.431848 & 0.011123 \\
0.727034 & 0.678398 & 1.124768 & -0.000008 & 1.15713 & 0.000 & 0.155846 & -0.00254 \\
0.177717 & 0.13062 & -0.09505 & -0.00013 & 0.01159 & 0.000 & 0.778338 & -0.00021 \\
0.301813 & 0.12316 & 0.271282 & -0.00018 & 0.03311 & 0.000 & 0.486024 & 0.011636 \\
0.414713 & -0.20925 & 1.222974 & 0.000054 & 0.10786 & 0.000 & 0.251595 & -0.01727 \\
1.700075 & 0.199318 & 0.225426 & -0.000032 & 0.01006 & 0.000 & 0.652782 & 0.030314 \\
-0.25459 & 0.132821 & 0.956079 & 0.000060 & 0.05553 & 0.000 & -0.37312 & -0.00054 \\
-2.25543 & -0.25148 & -0.92031 & -0.000032 & 0.07573 & 0.000 & 1.738408 & 0.027718 \\
0.318616 & 0.097084 & 0.13207 & -0.00012 & 0.16962 & 0.000 & 0.686528 & 0.01671 \\
0.680373 & 0.126166 & -0.27959 & -0.00014 & 0.04886 & 0.000 & 0.034872 & 0.012065 \\
\hline
\end{tabular}

\vspace{0.5cm}

\begin{tabular}{|rrrrrrrrr|}
\hline
Feature 9 & Feature 10 & Feature 11 & Feature 12 & Feature 13 & Feature 14 & Feature 15 & Feature 16 & Action Label \\
\hline
0.736442 & -0.26678 & 0.000 & -0.37522 & 0.034589 & 0.000 & 0.039715 & 0.060039 & 7 \\
-0.11608 & -1.53109 & 0.000 & -0.33782 & 0.695403 & 0.000 & -0.00257 & -0.01403 & 17 \\
0.492757 & -0.06871 & 0.000 & 1.107499 & 0.237301 & 0.000 & 0.008528 & 0.034149 & 7 \\
0.143643 & 0.03569 & 0.000 & 0.77755 & 0.199367 & 0.000 & 0.00316 & 0.029548 & 7 \\
0.018323 & -0.032362 & 0.000 & 1.897528 & -0.05634 & 0.000 & -0.00535 & 0.043167 & 2 \\
0.033747 & -0.42868 & 0.000 & 1.678755 & 0.028648 & 0.000 & 0.026876 & 0.021954 & 9 \\
-0.33624 & 0.12832 & 0.000 & -0.94617 & -0.06385 & 0.000 & 0.009777 & 0.102698 & 7 \\
-0.22714 & 0.32757 & 0.000 & 2.441481 & -0.37498 & 0.000 & 0.03634 & 0.068526 & 7 \\
0.168505 & -0.14289 & 0.000 & 0.88648 & 0.167437 & 0.000 & 0.00311 & 0.029505 & 7 \\
-0.32175 & -0.07368 & 0.000 & 1.56611 & -0.84533 & 0.000 & 0.000639 & 0.02832 & 7 \\
\hline
\end{tabular}
\caption{RoadRunner PPO boundary points: first 10 rows of features and action labels}
\label{tab:RoadRunner_PPO_boundary_points}
\end{table*}

\begin{table*}[!h]
\centering
\scriptsize
\begin{tabular}{|rrrrrrrr|}
\hline
Feature 1 & Feature 2 & Feature 3 & Feature 4 & Feature 5 & Feature 6 & Feature 7 & Feature 8 \\
\hline
393.7799 & 0.000 & 710.9127 & 0.000 & 0.000 & 0.000 & 706.2158 & 0.000 \\
0.000 & 16696.2 & 19701.97 & 0.000 & 0.000 & 0.000 & 1955.914 & 0.000 \\
1478.477 & 0.000 & 2021.436 & 0.000 & 0.000 & 0.000 & 727.3788 & 60.47329 \\
1949.78 & 0.000 & 440.0298 & 0.000 & 404.4525 & 0.000 & 883.1434 & 0.000 \\
1911.436 & 380.9045 & 242.8682 & 0.000 & 440.9319 & 0.000 & 1777.836 & 0.000 \\
2592.608 & 0.000 & 1427.355 & 0.000 & 0.000 & 0.000 & 1441.367 & 0.000 \\
1252.199 & 76.16316 & 441.7511 & 0.000 & 0.000 & 0.000 & 2203.14 & 0.000 \\
595.1301 & 804.1888 & 2465.322 & 0.000 & 0.000 & 0.000 & 657.2477 & 0.000 \\
1740.248 & 662.2636 & 1100.283 & 0.000 & 136.821 & 0.000 & 922.0039 & 0.000 \\
2394.39 & 496.4578 & 2170.893 & 0.000 & 20.7365 & 0.000 & 1717.301 & 0.000 \\
\hline
\end{tabular}

\vspace{0.5cm}

\begin{tabular}{|rrrrrrrrr|}
\hline
Feature 9 & Feature 10 & Feature 11 & Feature 12 & Feature 13 & Feature 14 & Feature 15 & Feature 16 & Action Label \\
\hline
135.8958 & 234.148 & 0.000 & 405.4395 & 74.99923 & 0.000 & 0.000 & 38.39702 & 7 \\
0.000 & 467.3831 & 0.000 & 2348.284 & 3777.795 & 0.000 & 0.000 & 0.000 & 17 \\
0.000 & 581.9508 & 0.000 & 1589.511 & 0.000 & 0.000 & 0.000 & 0.000 & 7 \\
511.5791 & 771.8015 & 0.000 & 1301.291 & 46.28678 & 0.000 & 63.36972 & 0.000 & 7 \\
1008.254 & 396.2003 & 0.000 & 1213.938 & 742.5882 & 0.000 & 0.000 & 0.000 & 2 \\
327.657 & 813.8804 & 0.000 & 308.6375 & 461.7389 & 0.000 & 0.000 & 0.000 & 9 \\
911.6214 & 491.1194 & 0.000 & 385.752 & 611.6022 & 0.000 & 0.000 & 0.000 & 7 \\
935.967 & 95.45456 & 0.000 & 1446.708 & 759.0374 & 0.000 & 0.000 & 0.000 & 7 \\
413.0294 & 505.3461 & 0.000 & 1325.548 & 217.6241 & 0.000 & 83.9481 & 0.000 & 7 \\
1186.176 & 474.8096 & 0.000 & 1641.388 & 1450.506 & 0.000 & 0.000 & 0.000 & 7 \\
\hline
\end{tabular}
\caption{RoadRunner PPO SHAP inverse boundary points: first 10 rows of features and action labels}
\label{tab:RoadRunner_PPO_shap_inverse_boundary_points}
\end{table*}

\clearpage
\section{Appendix F: Decision Trees} \label{Appendix F: Decision Trees}

\begin{figure}[htbp]
    \centering
    \includegraphics[width=\textwidth]{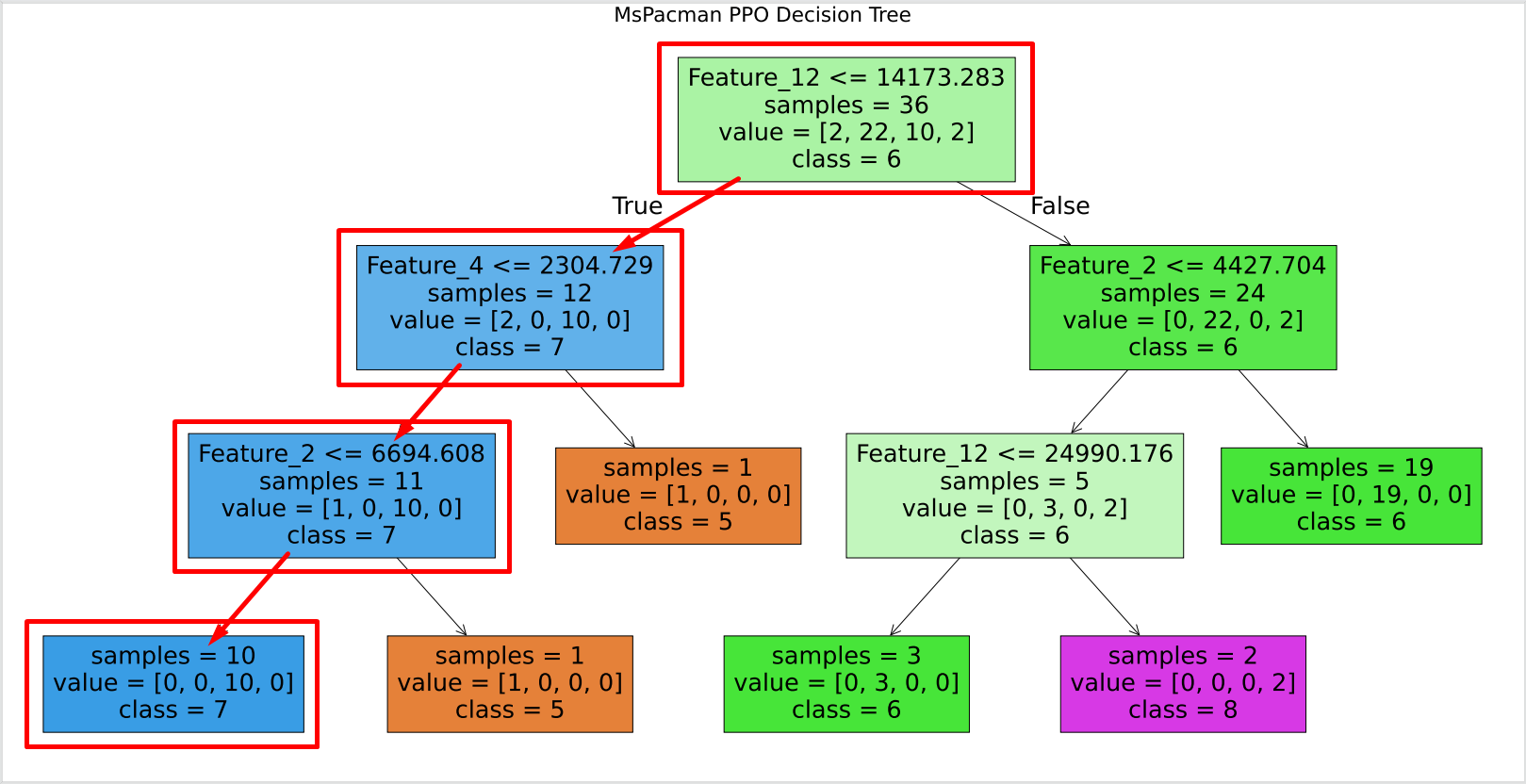}
    \caption{MsPacman PPO decision tree}
    \label{fig:MsPacman_ppo_decision_tree}
\end{figure}

\clearpage

\begin{figure}[htbp]
    \centering
    \includegraphics[width=\textwidth]{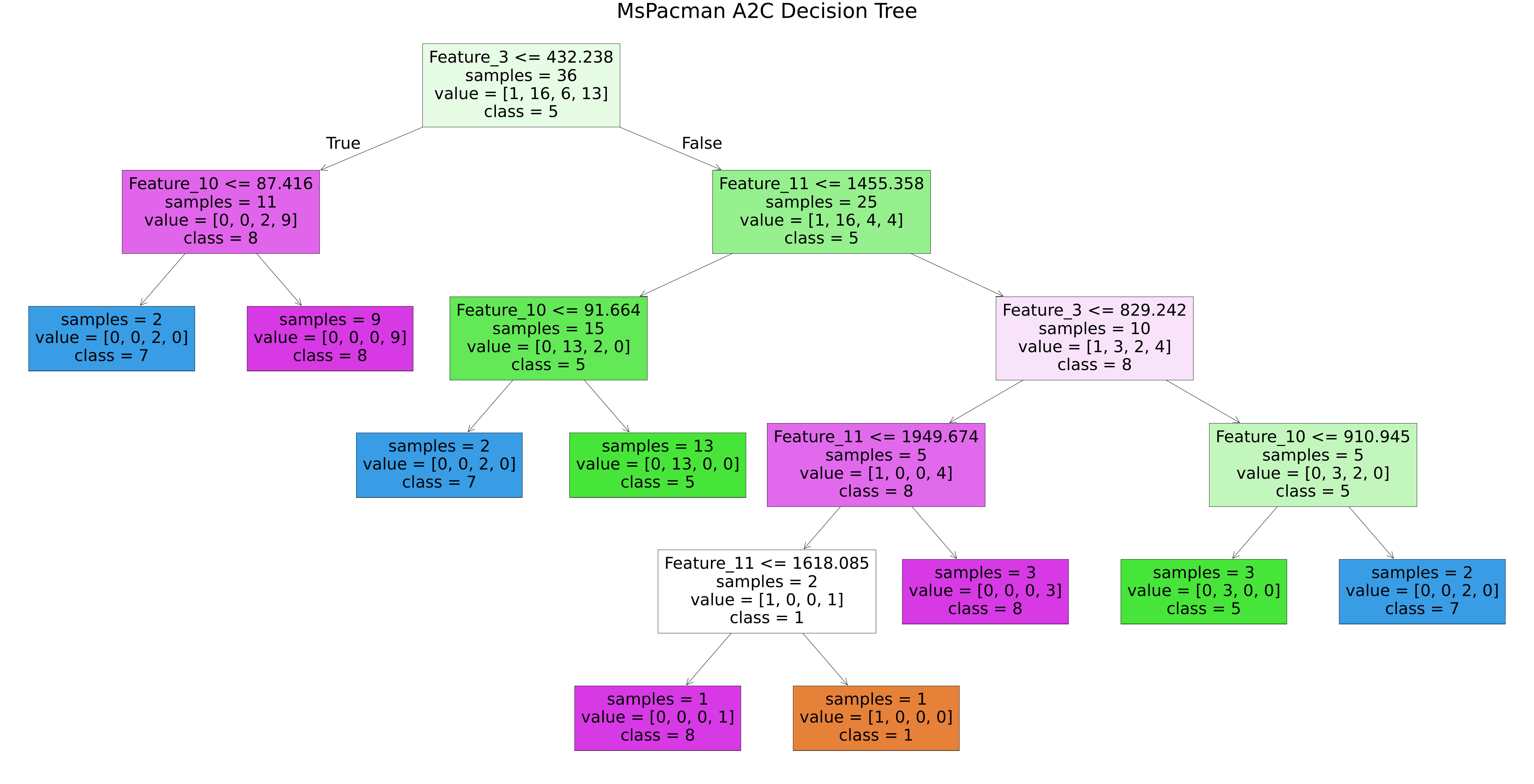}
    \caption{MsPacman A2C decision tree}
    \label{fig:MsPacman_a2c_decision_tree}
\end{figure}

\clearpage

\begin{figure}[htbp]
    \centering
    \includegraphics[width=\textwidth]{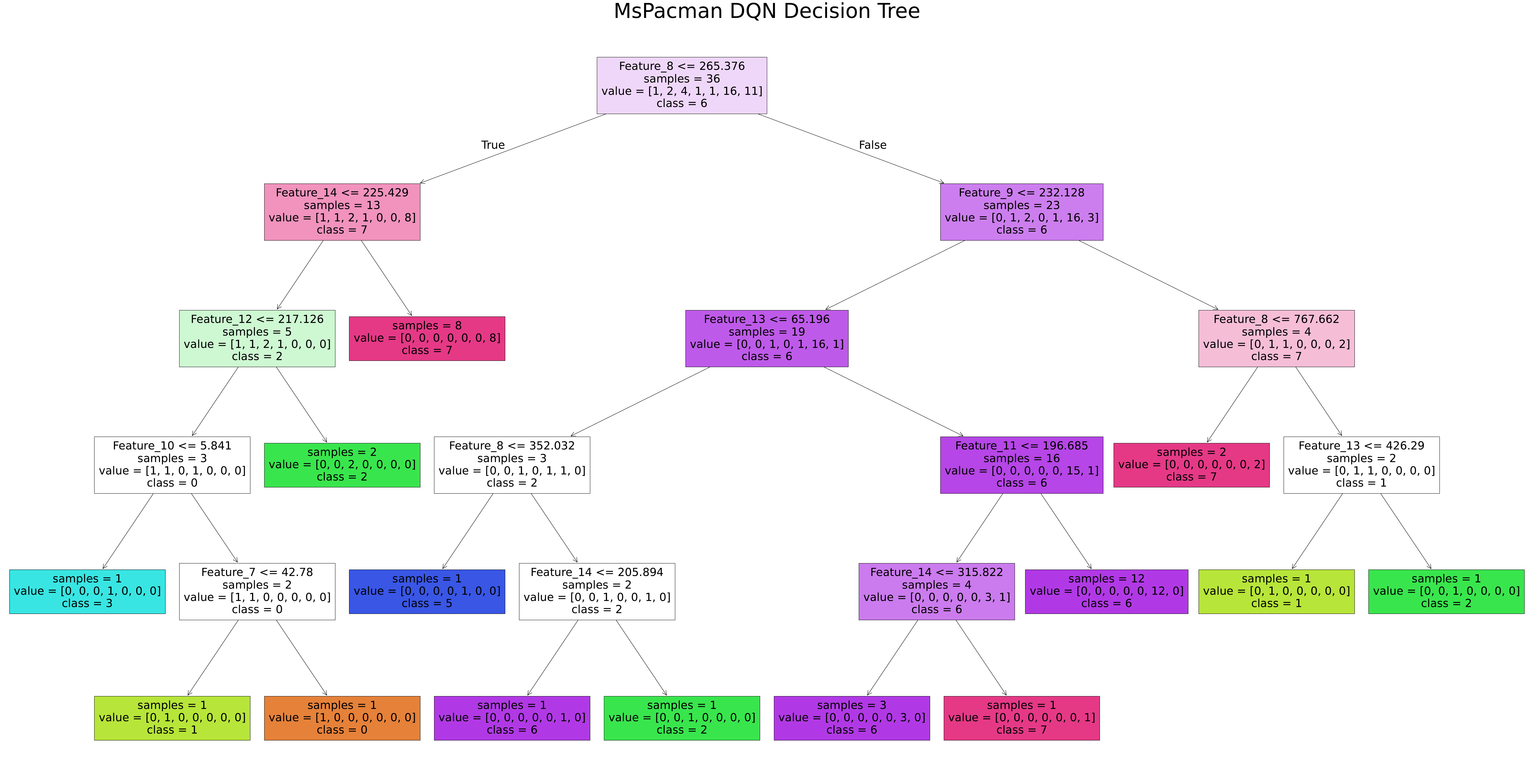}
    \caption{MsPacman DQN decision tree}
    \label{fig:MsPacman_dqn_decision_tree}
\end{figure}

\clearpage

\clearpage

\clearpage

\clearpage
\section{Appendix G: Linear Regression} \label{Appendix G: Linear Regression}

\begin{table}[htbp]
\centering
\renewcommand{\arraystretch}{1.3}
\begin{tabular}{|c|l|}
\hline
\textbf{Algorithm} & \textbf{Decision Boundary} \\ \hline
PPO & $
\begin{aligned}
f(x) &= 7.37 + 2.77 \times 10^{-4} \cdot x_{2} - 1.94 \times 10^{-4} \cdot x_{4} - 8.30 \times 10^{-3} \cdot x_{5} \\
     &\quad - 1.58 \times 10^{-3} \cdot x_{7} - 5.66 \times 10^{-4} \cdot x_{10} + 1.29 \times 10^{-3} \cdot x_{11} \\
     &\quad + 1.29 \times 10^{-5} \cdot x_{12} - 1.35 \times 10^{-4} \cdot x_{15}
\end{aligned}$ \\ \hline
A2C & $
\begin{aligned}
f(x) &= 4.95 + 4.05 \times 10^{-3} \cdot x_{3} - 7.39 \times 10^{-3} \cdot x_{8} + 4.76 \times 10^{-3} \cdot x_{9} \\
     &\quad - 6.18 \times 10^{-3} \cdot x_{10} + 3.51 \times 10^{-3} \cdot x_{11} + 3.47 \times 10^{-3} \cdot x_{14}
\end{aligned}$ \\ \hline
DQN & $
\begin{aligned}
f(x) &= 6.17 - 1.08 \times 10^{-3} \cdot x_{1} + 5.24 \times 10^{-2} \cdot x_{2} + 1.11 \times 10^{-3} \cdot x_{3} \\
     &\quad + 4.34 \times 10^{-3} \cdot x_{4} + 3.12 \times 10^{-3} \cdot x_{5} - 5.98 \times 10^{-2} \cdot x_{6} - 6.91 \times 10^{-2} \cdot x_{7} \\
     &\quad + 2.14 \times 10^{-3} \cdot x_{8} - 4.70 \times 10^{-3} \cdot x_{9} - 9.94 \times 10^{-4} \cdot x_{10} - 3.79 \times 10^{-3} \cdot x_{11} \\
     &\quad - 2.10 \times 10^{-3} \cdot x_{12} - 1.33 \times 10^{-3} \cdot x_{13} - 3.45 \times 10^{-5} \cdot x_{14} - 1.50 \times 10^{-3} \cdot x_{15} \\
     &\quad - 4.89 \times 10^{-3} \cdot x_{16}
\end{aligned}$ \\ \hline
\end{tabular}
\caption{MsPacman: interpretable policy boundary for linear regression}
\label{tab:MsPacman Linear Regression}
\end{table}

\begin{table}[htbp]
\centering
\renewcommand{\arraystretch}{1.3}
\begin{tabular}{|c|l|}
\hline
\textbf{Algorithm} & \textbf{Decision Boundary} \\ \hline
PPO & $
\begin{aligned}
f(x) &= 6.85 
+ 9.21 \times 10^{-4} \cdot x_{1} 
- 3.57 \times 10^{-4} \cdot x_{2} 
+ 1.14 \times 10^{-3} \cdot x_{3} 
- 4.51 \times 10^{-6} \cdot x_{5} \\
&\quad 
- 4.72 \times 10^{-4} \cdot x_{7} 
+ 1.59 \times 10^{-3} \cdot x_{8} 
+ 1.81 \times 10^{-4} \cdot x_{9} 
- 2.66 \times 10^{-4} \cdot x_{10} \\
&\quad 
- 2.73 \times 10^{-3} \cdot x_{12} 
- 2.05 \times 10^{-5} \cdot x_{13} 
+ 3.23 \times 10^{-3} \cdot x_{15} 
+ 2.21 \times 10^{-3} \cdot x_{16}
\end{aligned}$ \\ \hline
A2C & $
\begin{aligned}
f(x) &= 8.82 
- 9.00 \times 10^{-4} \cdot x_{1} 
- 5.92 \times 10^{-3} \cdot x_{2} 
+ 1.77 \times 10^{-3} \cdot x_{3} 
- 9.55 \times 10^{-3} \cdot x_{4} \\
&\quad 
- 6.94 \times 10^{-4} \cdot x_{5} 
- 4.30 \times 10^{-3} \cdot x_{6} 
+ 1.03 \times 10^{-4} \cdot x_{7} 
- 9.70 \times 10^{-4} \cdot x_{8} \\
&\quad 
+ 1.42 \times 10^{-2} \cdot x_{9} 
- 1.03 \times 10^{-3} \cdot x_{10} 
- 4.23 \times 10^{-3} \cdot x_{11} 
- 2.10 \times 10^{-3} \cdot x_{12} \\
&\quad 
- 1.02 \times 10^{-4} \cdot x_{13} 
- 4.19 \times 10^{-4} \cdot x_{14} 
+ 1.34 \times 10^{-3} \cdot x_{15} 
+ 1.08 \times 10^{-3} \cdot x_{16}
\end{aligned}$ \\ \hline
DQN & $
\begin{aligned}
f(x) &= 6.12 
+ 8.52 \times 10^{-3} \cdot x_{1} 
- 3.26 \times 10^{-2} \cdot x_{2} 
- 3.48 \times 10^{-3} \cdot x_{3} 
+ 4.69 \times 10^{-3} \cdot x_{4} \\
&\quad 
- 3.78 \times 10^{-3} \cdot x_{5} 
- 5.54 \times 10^{-2} \cdot x_{6} 
+ 2.44 \times 10^{-2} \cdot x_{7} 
- 2.91 \times 10^{-2} \cdot x_{8} \\
&\quad 
+ 3.47 \times 10^{-3} \cdot x_{9} 
+ 1.87 \times 10^{-2} \cdot x_{10} 
- 5.53 \times 10^{-3} \cdot x_{11} 
+ 6.60 \times 10^{-3} \cdot x_{12} \\
&\quad 
- 2.96 \times 10^{-3} \cdot x_{13} 
+ 1.75 \times 10^{-3} \cdot x_{14} 
- 4.66 \times 10^{-3} \cdot x_{15}
\end{aligned}$ \\ \hline
\end{tabular}
\caption{RoadRunner: interpretable policy boundary for linear regression}
\label{tab:RoadRunner Linear Regression}
\end{table}

\clearpage
\section{Appendix H: Logistic Regression} \label{Appendix H: Logistic Regression}

\begin{table}[htbp]
\centering
\begin{tabularx}{\textwidth}{c X}
\hline
Action $i$ & $f_i(x)$ \\
\hline
5 & $2.21\times 10^{-7} 
+ 9.61\times 10^{-4}x_{2} 
+ 1.69\times 10^{-4}x_{4} 
- 9.86\times 10^{-7}x_{5} 
+ 1.27\times 10^{-3}x_{7} 
+ 8.61\times 10^{-4}x_{10} 
+ 3.03\times 10^{-4}x_{11} 
- 4.15\times 10^{-4}x_{12} 
- 7.49\times 10^{-5}x_{15}$ \\

6 & $-1.14\times 10^{-7} 
- 2.17\times 10^{-3}x_{2} 
+ 2.28\times 10^{-3}x_{4} 
- 4.62\times 10^{-7}x_{5} 
- 8.96\times 10^{-4}x_{7} 
+ 1.08\times 10^{-3}x_{10} 
- 2.40\times 10^{-4}x_{11} 
+ 1.17\times 10^{-3}x_{12} 
- 4.35\times 10^{-4}x_{15}$ \\

7 & $6.06\times 10^{-7} 
+ 1.90\times 10^{-3}x_{2} 
- 1.47\times 10^{-3}x_{4} 
+ 1.58\times 10^{-6}x_{5} 
- 3.68\times 10^{-4}x_{7} 
+ 1.11\times 10^{-3}x_{10} 
- 5.90\times 10^{-5}x_{11} 
- 2.23\times 10^{-3}x_{12} 
+ 1.75\times 10^{-3}x_{15}$ \\

8 & $-7.12\times 10^{-7} 
- 6.93\times 10^{-4}x_{2} 
- 9.76\times 10^{-4}x_{4} 
- 1.30\times 10^{-7}x_{5} 
- 9.69\times 10^{-6}x_{7} 
- 3.05\times 10^{-3}x_{10} 
- 4.47\times 10^{-6}x_{11} 
+ 1.47\times 10^{-3}x_{12} 
- 1.23\times 10^{-3}x_{15}$ \\
\hline
\end{tabularx}
\caption{MsPacman PPO: interpretable policy boundary for logistic regression}
\label{tab:MsPacman_PPO_LogReg}
\end{table}

\begin{table}[htbp]
\centering
\begin{tabularx}{\textwidth}{c X}
\hline
Action $i$ & $f_i(x)$ \\
\hline
1 & $-4.54\times 10^{-5}
- 6.57\times 10^{-3}x_{3}
+ 8.94\times 10^{-3}x_{8}
- 8.20\times 10^{-5}x_{9}
+ 6.73\times 10^{-3}x_{10}
- 4.47\times 10^{-3}x_{11}
+ 6.11\times 10^{-3}x_{14}$ \\

5 & $6.59\times 10^{-5}
+ 9.31\times 10^{-3}x_{3}
+ 2.13\times 10^{-2}x_{8}
- 7.33\times 10^{-3}x_{9}
+ 3.39\times 10^{-3}x_{10}
- 1.33\times 10^{-2}x_{11}
- 6.96\times 10^{-3}x_{14}$ \\

7 & $-2.49\times 10^{-5}
+ 1.62\times 10^{-2}x_{3}
- 1.46\times 10^{-2}x_{8}
+ 7.45\times 10^{-3}x_{9}
- 5.94\times 10^{-3}x_{10}
+ 1.26\times 10^{-3}x_{11}
+ 6.96\times 10^{-4}x_{14}$ \\

8 & $4.39\times 10^{-6}
- 1.89\times 10^{-2}x_{3}
- 1.57\times 10^{-2}x_{8}
- 3.56\times 10^{-5}x_{9}
- 4.18\times 10^{-3}x_{10}
+ 1.66\times 10^{-2}x_{11}
+ 1.57\times 10^{-4}x_{14}$ \\
\hline
\end{tabularx}
\caption{MsPacman A2C: interpretable policy boundary for logistic regression}
\label{tab:MsPacman_A2C_LogReg}
\end{table}

\begin{table*}[htbp]
\centering
\begin{tabularx}{\textwidth}{c X}
\hline
Action $i$ & $f_i(x)$ \\
\hline
0 & $9.37\times 10^{-5} - 2.29\times 10^{-4}x_{1} - 7.40\times 10^{-5}x_{2} - 7.46\times 10^{-5}x_{3} - 1.48\times 10^{-3}x_{4} - 1.61\times 10^{-4}x_{5} - 3.56\times 10^{-4}x_{6} + 2.29\times 10^{-2}x_{7} - 1.61\times 10^{-2}x_{8} - 3.53\times 10^{-3}x_{9} - 8.99\times 10^{-3}x_{10} - 4.98\times 10^{-3}x_{11} + 4.51\times 10^{-3}x_{12} - 3.42\times 10^{-3}x_{13} + 1.51\times 10^{-2}x_{14} - 5.62\times 10^{-5}x_{15} - 1.03\times 10^{-3}x_{16}$ \\

1 & $1.50\times 10^{-5} - 2.97\times 10^{-4}x_{1} - 9.49\times 10^{-4}x_{2} - 2.25\times 10^{-4}x_{3} - 5.42\times 10^{-3}x_{4} - 2.22\times 10^{-3}x_{5} + 9.06\times 10^{-3}x_{6} - 3.71\times 10^{-4}x_{7} + 8.84\times 10^{-3}x_{8} + 2.66\times 10^{-2}x_{9} + 7.08\times 10^{-3}x_{10} - 7.55\times 10^{-3}x_{11} - 1.09\times 10^{-2}x_{12} + 4.77\times 10^{-5}x_{13} - 1.05\times 10^{-2}x_{14} - 1.50\times 10^{-2}x_{15} - 1.34\times 10^{-3}x_{16}$ \\

2 & $-1.06\times 10^{-4} + 1.05\times 10^{-3}x_{1} - 1.10\times 10^{-4}x_{2} - 2.38\times 10^{-4}x_{3} - 2.91\times 10^{-3}x_{4} - 5.65\times 10^{-3}x_{5} + 1.32\times 10^{-4}x_{6} - 5.77\times 10^{-3}x_{7} - 1.29\times 10^{-2}x_{8} - 2.81\times 10^{-3}x_{9} + 8.72\times 10^{-3}x_{10} + 2.47\times 10^{-2}x_{11} + 1.65\times 10^{-2}x_{12} - 9.90\times 10^{-3}x_{13} + 1.78\times 10^{-3}x_{14} + 2.02\times 10^{-2}x_{15} + 4.76\times 10^{-3}x_{16}$ \\

3 & $-2.32\times 10^{-5} - 9.31\times 10^{-5}x_{1} - 8.10\times 10^{-5}x_{2} + 1.04\times 10^{-2}x_{3} - 1.94\times 10^{-3}x_{4} + 4.20\times 10^{-3}x_{5} - 4.29\times 10^{-4}x_{6} - 3.89\times 10^{-4}x_{7} - 8.58\times 10^{-3}x_{8} - 2.01\times 10^{-3}x_{9} - 1.65\times 10^{-3}x_{10} + 8.15\times 10^{-3}x_{11} - 3.12\times 10^{-3}x_{12} + 7.88\times 10^{-3}x_{13} - 7.57\times 10^{-3}x_{14} - 1.10\times 10^{-4}x_{15} - 4.20\times 10^{-4}x_{16}$ \\

5 & $1.50\times 10^{-4} - 6.85\times 10^{-5}x_{1} - 2.54\times 10^{-4}x_{2} - 1.30\times 10^{-4}x_{3} - 4.34\times 10^{-3}x_{4} - 3.80\times 10^{-3}x_{5} - 6.65\times 10^{-4}x_{6} - 1.06\times 10^{-3}x_{7} + 3.67\times 10^{-3}x_{8} - 1.07\times 10^{-2}x_{9} - 3.67\times 10^{-3}x_{10} + 4.52\times 10^{-3}x_{11} - 6.73\times 10^{-3}x_{12} + 5.65\times 10^{-3}x_{13} + 1.72\times 10^{-3}x_{14} - 5.14\times 10^{-3}x_{15} - 3.09\times 10^{-4}x_{16}$ \\

6 & $-8.05\times 10^{-5} - 7.67\times 10^{-5}x_{1} + 9.98\times 10^{-3}x_{2} - 9.69\times 10^{-3}x_{3} + 6.58\times 10^{-3}x_{4} + 3.64\times 10^{-3}x_{5} + 4.22\times 10^{-3}x_{6} - 2.66\times 10^{-4}x_{7} + 7.39\times 10^{-3}x_{8} - 1.75\times 10^{-2}x_{9} - 2.85\times 10^{-4}x_{10} + 8.63\times 10^{-3}x_{11} + 2.29\times 10^{-3}x_{12} + 1.94\times 10^{-2}x_{13} - 2.01\times 10^{-2}x_{14} - 1.13\times 10^{-2}x_{15} - 3.46\times 10^{-4}x_{16}$ \\

7 & $-4.95\times 10^{-5} - 2.91\times 10^{-4}x_{1} - 8.51\times 10^{-3}x_{2} - 6.61\times 10^{-5}x_{3} + 9.50\times 10^{-3}x_{4} + 3.99\times 10^{-3}x_{5} - 1.20\times 10^{-2}x_{6} - 1.51\times 10^{-2}x_{7} + 1.77\times 10^{-2}x_{8} + 9.91\times 10^{-3}x_{9} - 1.22\times 10^{-3}x_{10} - 3.35\times 10^{-2}x_{11} - 2.57\times 10^{-3}x_{12} - 1.97\times 10^{-2}x_{13} + 1.95\times 10^{-2}x_{14} + 1.14\times 10^{-2}x_{15} - 1.31\times 10^{-3}x_{16}$ \\
\hline
\end{tabularx}
\caption{MsPacman DQN: interpretable policy boundary for logistic regression}
\label{tab:MsPacman_DQN_LogReg}
\end{table*}

\begin{table*}[htbp]
\centering
\begin{tabularx}{\textwidth}{c X}
\hline
Action $i$ & $f_i(x)$ \\
\hline
2 & $-5.32\times 10^{-6}
- 2.45\times 10^{-3}x_{1}
- 6.40\times 10^{-4}x_{2}
- 4.97\times 10^{-3}x_{3}
- 2.69\times 10^{-4}x_{5}
+ 2.25\times 10^{-3}x_{7}
- 2.26\times 10^{-7}x_{8}
- 4.24\times 10^{-4}x_{9}
- 8.58\times 10^{-4}x_{10}
+ 9.71\times 10^{-3}x_{12}
+ 1.69\times 10^{-3}x_{13}
+ 3.58\times 10^{-4}x_{15}
- 1.90\times 10^{-4}x_{16}$ \\

7 & $6.16\times 10^{-6}
+ 2.72\times 10^{-3}x_{1}
+ 5.40\times 10^{-4}x_{2}
- 1.87\times 10^{-4}x_{3}
- 8.33\times 10^{-6}x_{5}
+ 5.53\times 10^{-4}x_{7}
- 3.31\times 10^{-4}x_{8}
- 9.47\times 10^{-4}x_{9}
+ 8.05\times 10^{-4}x_{10}
+ 1.88\times 10^{-3}x_{12}
- 5.16\times 10^{-4}x_{13}
- 4.61\times 10^{-4}x_{15}
- 5.13\times 10^{-4}x_{16}$ \\

9 & $2.92\times 10^{-6}
+ 5.07\times 10^{-3}x_{1}
- 1.63\times 10^{-3}x_{2}
+ 3.06\times 10^{-3}x_{3}
+ 5.36\times 10^{-4}x_{5}
- 3.27\times 10^{-3}x_{7}
+ 6.30\times 10^{-4}x_{8}
+ 2.57\times 10^{-3}x_{9}
+ 1.44\times 10^{-3}x_{10}
- 8.63\times 10^{-3}x_{12}
- 1.69\times 10^{-3}x_{13}
+ 1.53\times 10^{-4}x_{15}
+ 4.53\times 10^{-5}x_{16}$ \\

12 & $-1.73\times 10^{-6}
- 3.51\times 10^{-3}x_{1}
- 1.48\times 10^{-4}x_{2}
+ 1.69\times 10^{-3}x_{3}
- 1.99\times 10^{-4}x_{5}
+ 2.07\times 10^{-3}x_{7}
- 2.59\times 10^{-4}x_{8}
- 3.22\times 10^{-4}x_{9}
- 1.08\times 10^{-3}x_{10}
- 2.16\times 10^{-3}x_{12}
+ 5.44\times 10^{-4}x_{13}
- 3.16\times 10^{-5}x_{15}
+ 7.63\times 10^{-4}x_{16}$ \\

17 & $-2.03\times 10^{-6}
- 1.83\times 10^{-3}x_{1}
+ 1.88\times 10^{-3}x_{2}
+ 4.05\times 10^{-4}x_{3}
- 5.99\times 10^{-5}x_{5}
- 1.60\times 10^{-3}x_{7}
- 3.94\times 10^{-5}x_{8}
- 8.73\times 10^{-4}x_{9}
- 3.05\times 10^{-4}x_{10}
- 8.01\times 10^{-4}x_{12}
- 1.89\times 10^{-5}x_{13}
- 1.76\times 10^{-5}x_{15}
- 1.06\times 10^{-4}x_{16}$ \\
\hline
\end{tabularx}
\caption{RoadRunner PPO: interpretable policy boundary for logistic regression}
\label{tab:RoadRunne_PPO_LogReg}
\end{table*}

\begin{table*}[htbp]
\centering
\begin{tabularx}{\textwidth}{c X}
\hline
Action $i$ & $f_i(x)$ \\
\hline
2 & $-2.86\times 10^{-6}
- 1.35\times 10^{-3}x_{1}
+ 1.61\times 10^{-3}x_{2}
- 1.98\times 10^{-5}x_{3}
+ 1.56\times 10^{-3}x_{4}
- 1.17\times 10^{-2}x_{5}
+ 3.76\times 10^{-3}x_{6}
- 6.08\times 10^{-4}x_{7}
+ 1.83\times 10^{-2}x_{8}
- 3.67\times 10^{-6}x_{9}
- 1.45\times 10^{-3}x_{10}
+ 2.28\times 10^{-3}x_{11}
+ 8.55\times 10^{-3}x_{12}
+ 6.14\times 10^{-3}x_{13}
- 2.22\times 10^{-3}x_{14}
- 2.16\times 10^{-3}x_{15}
- 8.77\times 10^{-3}x_{16}$ \\

4 & $-5.35\times 10^{-5}
+ 3.34\times 10^{-3}x_{1}
- 1.96\times 10^{-3}x_{2}
- 6.08\times 10^{-4}x_{3}
- 6.50\times 10^{-5}x_{4}
+ 4.06\times 10^{-3}x_{5}
- 1.04\times 10^{-3}x_{6}
+ 2.62\times 10^{-3}x_{7}
- 5.72\times 10^{-3}x_{8}
- 7.24\times 10^{-6}x_{9}
+ 3.40\times 10^{-3}x_{10}
- 3.75\times 10^{-5}x_{11}
- 4.91\times 10^{-3}x_{12}
- 2.44\times 10^{-3}x_{13}
- 6.74\times 10^{-3}x_{14}
- 1.03\times 10^{-3}x_{15}
+ 3.41\times 10^{-3}x_{16}$ \\

5 & $-2.82\times 10^{-5}
- 7.23\times 10^{-3}x_{1}
+ 1.14\times 10^{-2}x_{2}
- 1.97\times 10^{-5}x_{3}
- 1.84\times 10^{-5}x_{4}
- 6.60\times 10^{-3}x_{5}
- 1.41\times 10^{-3}x_{6}
- 6.26\times 10^{-3}x_{7}
- 8.87\times 10^{-3}x_{8}
+ 2.14\times 10^{-5}x_{9}
- 7.41\times 10^{-3}x_{10}
- 5.33\times 10^{-5}x_{11}
- 3.18\times 10^{-3}x_{12}
+ 7.26\times 10^{-3}x_{13}
+ 1.81\times 10^{-3}x_{14}
+ 2.15\times 10^{-3}x_{15}
+ 3.50\times 10^{-3}x_{16}$ \\

7 & $-1.50\times 10^{-4}
- 2.88\times 10^{-3}x_{1}
+ 2.75\times 10^{-3}x_{2}
- 7.10\times 10^{-4}x_{3}
- 1.06\times 10^{-3}x_{4}
+ 1.15\times 10^{-2}x_{5}
+ 1.82\times 10^{-3}x_{6}
+ 8.55\times 10^{-3}x_{7}
+ 1.14\times 10^{-2}x_{8}
- 4.94\times 10^{-5}x_{9}
+ 6.46\times 10^{-3}x_{10}
- 1.96\times 10^{-3}x_{11}
+ 1.15\times 10^{-2}x_{12}
+ 1.57\times 10^{-3}x_{13}
- 3.53\times 10^{-3}x_{14}
+ 1.96\times 10^{-3}x_{15}
- 7.76\times 10^{-3}x_{16}$ \\

9 & $2.34\times 10^{-4}
+ 8.11\times 10^{-3}x_{1}
- 1.38\times 10^{-2}x_{2}
+ 1.36\times 10^{-3}x_{3}
- 4.17\times 10^{-4}x_{4}
+ 2.76\times 10^{-3}x_{5}
- 3.13\times 10^{-3}x_{6}
- 4.31\times 10^{-3}x_{7}
- 1.51\times 10^{-2}x_{8}
+ 3.89\times 10^{-5}x_{9}
- 9.97\times 10^{-4}x_{10}
- 2.26\times 10^{-4}x_{11}
- 1.20\times 10^{-2}x_{12}
- 1.25\times 10^{-2}x_{13}
+ 1.07\times 10^{-2}x_{14}
- 9.17\times 10^{-4}x_{15}
+ 9.62\times 10^{-3}x_{16}$ \\
\hline
\end{tabularx}
\caption{RoadRunner A2C: interpretable policy boundary for logistic regression}
\label{tab:RoadRunne_A2C_LogReg}
\end{table*}

\begin{table*}[htbp]
\centering
\begin{tabularx}{\textwidth}{c X}
\hline
Action $i$ & $f_i(x)$ \\
\hline
2 & $-3.21\times 10^{-4}
- 2.84\times 10^{-3}x_{1}
+ 6.14\times 10^{-3}x_{2}
+ 2.54\times 10^{-3}x_{3}
- 5.52\times 10^{-4}x_{4}
- 1.67\times 10^{-2}x_{5}
+ 9.15\times 10^{-3}x_{6}
+ 7.05\times 10^{-2}x_{7}
+ 1.98\times 10^{-2}x_{8}
- 3.41\times 10^{-3}x_{9}
- 1.92\times 10^{-2}x_{10}
- 3.83\times 10^{-3}x_{11}
- 3.61\times 10^{-2}x_{12}
+ 3.10\times 10^{-2}x_{13}
- 7.27\times 10^{-3}x_{14}
+ 1.95\times 10^{-2}x_{15}$ \\

3 & $3.29\times 10^{-3}
- 2.14\times 10^{-2}x_{1}
+ 4.69\times 10^{-2}x_{2}
+ 1.04\times 10^{-2}x_{3}
- 4.94\times 10^{-3}x_{4}
+ 1.05\times 10^{-2}x_{5}
+ 1.40\times 10^{-2}x_{6}
+ 9.10\times 10^{-3}x_{7}
+ 4.96\times 10^{-3}x_{8}
+ 1.89\times 10^{-2}x_{9}
- 1.79\times 10^{-3}x_{10}
- 6.09\times 10^{-3}x_{11}
+ 2.87\times 10^{-2}x_{12}
+ 1.80\times 10^{-2}x_{13}
- 8.07\times 10^{-3}x_{14}
- 4.53\times 10^{-2}x_{15}$ \\

5 & $-3.82\times 10^{-3}
+ 1.38\times 10^{-2}x_{1}
- 2.11\times 10^{-2}x_{2}
+ 5.82\times 10^{-3}x_{3}
- 4.63\times 10^{-3}x_{4}
+ 3.23\times 10^{-2}x_{5}
- 7.31\times 10^{-3}x_{6}
- 3.31\times 10^{-2}x_{7}
+ 1.94\times 10^{-2}x_{8}
+ 7.67\times 10^{-3}x_{9}
- 2.10\times 10^{-2}x_{10}
+ 2.40\times 10^{-2}x_{11}
+ 3.09\times 10^{-2}x_{12}
- 1.02\times 10^{-2}x_{13}
- 6.90\times 10^{-3}x_{14}
+ 2.54\times 10^{-2}x_{15}$ \\

6 & $-2.42\times 10^{-4}
+ 5.95\times 10^{-4}x_{1}
- 2.69\times 10^{-3}x_{2}
- 1.15\times 10^{-2}x_{3}
- 5.70\times 10^{-4}x_{4}
- 1.82\times 10^{-2}x_{5}
- 2.29\times 10^{-3}x_{6}
- 1.28\times 10^{-2}x_{7}
+ 1.02\times 10^{-2}x_{8}
+ 6.67\times 10^{-3}x_{9}
+ 3.24\times 10^{-2}x_{10}
- 8.66\times 10^{-4}x_{11}
- 4.73\times 10^{-2}x_{12}
- 2.27\times 10^{-2}x_{13}
+ 2.38\times 10^{-3}x_{14}
+ 9.63\times 10^{-2}x_{15}$ \\

7 & $-1.45\times 10^{-3}
- 1.09\times 10^{-3}x_{1}
- 1.09\times 10^{-2}x_{2}
- 1.22\times 10^{-2}x_{3}
- 1.40\times 10^{-3}x_{4}
- 3.52\times 10^{-2}x_{5}
- 8.33\times 10^{-3}x_{6}
- 4.28\times 10^{-2}x_{7}
+ 3.76\times 10^{-2}x_{8}
- 3.87\times 10^{-3}x_{9}
+ 2.25\times 10^{-2}x_{10}
- 2.47\times 10^{-3}x_{11}
+ 1.18\times 10^{-2}x_{12}
- 5.57\times 10^{-2}x_{13}
- 1.46\times 10^{-2}x_{14}
+ 4.74\times 10^{-2}x_{15}$ \\

8 & $2.05\times 10^{-3}
+ 1.61\times 10^{-2}x_{1}
- 2.26\times 10^{-2}x_{2}
- 3.22\times 10^{-2}x_{3}
+ 1.28\times 10^{-2}x_{4}
+ 2.20\times 10^{-2}x_{5}
- 9.96\times 10^{-5}x_{6}
- 9.35\times 10^{-3}x_{7}
- 6.44\times 10^{-2}x_{8}
+ 2.68\times 10^{-3}x_{9}
- 4.21\times 10^{-2}x_{10}
- 2.70\times 10^{-3}x_{11}
+ 6.59\times 10^{-2}x_{12}
+ 2.59\times 10^{-2}x_{13}
+ 3.06\times 10^{-2}x_{14}
- 5.24\times 10^{-2}x_{15}$ \\

9 & $1.85\times 10^{-4}
- 1.12\times 10^{-3}x_{1}
+ 8.45\times 10^{-3}x_{2}
+ 6.73\times 10^{-2}x_{3}
- 1.43\times 10^{-4}x_{4}
- 3.34\times 10^{-2}x_{5}
- 3.35\times 10^{-3}x_{6}
+ 2.18\times 10^{-2}x_{7}
- 6.23\times 10^{-3}x_{8}
- 2.46\times 10^{-2}x_{9}
+ 1.36\times 10^{-2}x_{10}
- 2.80\times 10^{-3}x_{11}
+ 6.52\times 10^{-3}x_{12}
- 3.59\times 10^{-2}x_{13}
+ 1.21\times 10^{-2}x_{14}
- 1.08\times 10^{-1}x_{15}$ \\

11 & $1.27\times 10^{-4}
- 2.76\times 10^{-3}x_{1}
- 2.73\times 10^{-3}x_{2}
- 1.22\times 10^{-2}x_{3}
- 2.85\times 10^{-4}x_{4}
+ 4.62\times 10^{-2}x_{5}
- 1.06\times 10^{-3}x_{6}
- 2.13\times 10^{-3}x_{7}
- 9.06\times 10^{-3}x_{8}
- 2.45\times 10^{-3}x_{9}
- 1.12\times 10^{-3}x_{10}
- 2.52\times 10^{-3}x_{11}
- 1.54\times 10^{-2}x_{12}
+ 2.54\times 10^{-2}x_{13}
- 4.69\times 10^{-3}x_{14}
- 1.96\times 10^{-2}x_{15}$ \\

15 & $1.85\times 10^{-4}
- 1.27\times 10^{-3}x_{1}
- 1.34\times 10^{-3}x_{2}
- 1.79\times 10^{-2}x_{3}
- 2.49\times 10^{-4}x_{4}
- 7.45\times 10^{-3}x_{5}
- 6.78\times 10^{-4}x_{6}
- 1.20\times 10^{-3}x_{7}
- 1.23\times 10^{-2}x_{8}
- 1.51\times 10^{-3}x_{9}
+ 1.67\times 10^{-2}x_{10}
- 2.71\times 10^{-3}x_{11}
- 4.50\times 10^{-2}x_{12}
+ 2.42\times 10^{-2}x_{13}
- 3.47\times 10^{-3}x_{14}
+ 3.71\times 10^{-2}x_{15}$ \\
\hline
\end{tabularx}
\caption{RoadRunner DQN: interpretable policy boundary for logistic regression}
\label{tab:RoadRunner_DQN_LogReg}
\end{table*}

\end{document}